%% file: main.tex
\documentclass{article}

\usepackage{microtype}
\usepackage{graphicx}
\usepackage{subcaption}
\usepackage{booktabs} % for professional tables

\usepackage{hyperref}

\usepackage[preprint]{icml2026}

\usepackage{amsmath}
\usepackage{amssymb}
\usepackage{mathtools}
\usepackage{amsthm}

\usepackage[capitalize,noabbrev]{cleveref}

\theoremstyle{plain}

\theoremstyle{definition}

\theoremstyle{remark}

\usepackage[textsize=tiny]{todonotes}

\usepackage{titletoc}

\usepackage{algorithm}
\usepackage{algorithmic}

\usepackage{multirow}

\usepackage{xcolor}
\definecolor{mydarkblue}{rgb}{0,0.08,0.45}
\definecolor{citecolor}{RGB}{21, 152, 56} % 绿色

\usepackage{amsmath}
\usepackage{amssymb}
\usepackage{mathtools}
\usepackage{amsthm}

\usepackage{pifont}% http://ctan.org/pkg/pifont

\definecolor{graybg}{gray}{.92}
\definecolor{bgcolor}{RGB}{246,248,250}

\newcommand\bench{WorldGUI}
\newcommand\agent{WorldGUI-Agent}

\usepackage{colortbl}
\usepackage[dvipsnames]{xcolor}
\newcommand{\cmark}{\color{OliveGreen}{\ding{51}}}%
\newcommand{\xmark}{\color{Maroon}{\ding{55}}}%

\icmltitlerunning{Submission and Formatting Instructions for ICML 2026}

\begin{document}

\twocolumn[
  \icmltitle{WorldGUI: An Interactive Benchmark for Desktop GUI Automation from Any Starting Point}



  \icmlsetsymbol{corr}{$\dagger$}

  \begin{icmlauthorlist}
    \icmlauthor{Henry Hengyuan Zhao}{1}
    \icmlauthor{Kaiming Yang}{1}
    \icmlauthor{Wendi Yu}{1}
    \icmlauthor{Difei Gao}{1}
    \icmlauthor{Mike Zheng Shou}{1,corr}
  \end{icmlauthorlist}

  \icmlaffiliation{1}{Show Lab, National University of Singapore}

  \icmlcorrespondingauthor{Mike Zheng Shou}{mike.zheng.shou@gmail.com}


  \vskip 0.3in
]



\printAffiliationsAndNotice{}  % no special notice (required even if empty)

\input{sections/1_abs}

\input{sections/2_intro}

\input{sections/8_related}

\input{sections/3_bench}

\input{sections/5_exp}

\input{sections/6_con}

\nocite{langley00}

\bibliography{main}
\bibliographystyle{icml2026}

%%%%%%%%%%%%%%%%%%%%%%%%%%%%%%%%%%%%%%%%%%%%%%%%%%%%%%%%%%%%%%%%%%%%%%%%%%%%%%%
%%%%%%%%%%%%%%%%%%%%%%%%%%%%%%%%%%%%%%%%%%%%%%%%%%%%%%%%%%%%%%%%%%%%%%%%%%%%%%%
% APPENDIX
%%%%%%%%%%%%%%%%%%%%%%%%%%%%%%%%%%%%%%%%%%%%%%%%%%%%%%%%%%%%%%%%%%%%%%%%%%%%%%%
%%%%%%%%%%%%%%%%%%%%%%%%%%%%%%%%%%%%%%%%%%%%%%%%%%%%%%%%%%%%%%%%%%%%%%%%%%%%%%%

\clearpage

\startcontents[app]
% % 第2个参数 l = 左对齐风格；第3个参数 1 = 目录深度（1=section，2=subsection，依需要调）
\printcontents[app]{l}{1}{}

\newpage
\appendix

\input{sections/7_append}

%%%%%%%%%%%%%%%%%%%%%%%%%%%%%%%%%%%%%%%%%%%%%%%%%%%%%%%%%%%%%%%%%%%%%%%%%%%%%%%
%%%%%%%%%%%%%%%%%%%%%%%%%%%%%%%%%%%%%%%%%%%%%%%%%%%%%%%%%%%%%%%%%%%%%%%%%%%%%%%

\end{document}

%% file: sections/1_abs.tex
% \begin{abstract}

% GUI agents have achieved outstanding performance in GUI element grounding. However, planning remains highly challenging, particularly due to sensitivity to the initial state of the environment. Even slight differences in the initial state, such as the target software not being open or the interface not being in its default configuration, can lead to planning failures. This issue is pervasive in real-world application scenarios, yet existing benchmarks fail to adequately evaluate it.
% To address this gap, we introduce \textbf{\bench{}}, a comprehensive GUI benchmark comprising tasks across ten widely used desktop and web applications, such as PowerPoint, VSCode, and Acrobat. Each task is instantiated under diverse initial states to simulate authentic human–computer interactions. In addition, we propose \textbf{\agent{}}, a simple and general framework that unifies three critic-based modules as the alongside baseline concentrated on the enviroment state changes.
% Experimental results show that existing agents still struggle on desktop GUI tasks and exhibit significant performance discrepancy between default and non-default initial conditions. These findings highlight their limited robustness and fragile planning capabilities.

% \end{abstract}

\begin{abstract}
Recent progress in GUI agents has substantially improved visual grounding, yet robust planning remains challenging, particularly when the environment deviates from a canonical initial state. 
In real applications, users often invoke assistance mid-workflow, where software may be partially configured, steps may have been executed in different orders, or the interface may differ from its default setup. 
Such task-state variability is pervasive but insufficiently evaluated in existing GUI benchmarks. To address this gap, we introduce \textbf{\bench{}}, a benchmark covering ten widely used desktop and web applications with tasks instantiated under diverse, systematically constructed initial states. 
These variations capture realistic human–computer interaction settings and enable diagnostic evaluation of an agent’s ability to recover, adapt plans, and handle non-default contexts. We further present \textbf{\agent{}}, a simple and model-agnostic framework that organizes planning and execution around three critique stages, improving reliability in dynamic environments. Experiments demonstrate that state-of-the-art GUI agents exhibit substantial performance degradation under non-default initial conditions, revealing limited robustness and fragile planning behaviors. 
Our benchmark and framework provide a foundation for developing more adaptable and reliable GUI agents. The code and data are available at \href{https://github.com/showlab/WorldGUI}{https://github.com/showlab/WorldGUI}.
\end{abstract}

%% file: sections/2_intro.tex
\begin{table*}[t]
    % \vskip -0.2in
    \caption{\textbf{Comparison with other online GUI benchmarks.} \textbf{\bench{}} is a unique benchmark that embraces diverse initial states and better reflects the authentic interactions in GUI scenarios. Env?: Indicates whether an environment is required to be deployed.}
    \label{tab:benchmarkcompare}
    \vskip -0.1in
    \centering
    \resizebox{0.8\linewidth}{!}{
    \renewcommand\arraystretch{1}
    \tabcolsep=0.5mm
    \begin{tabular}{lccccccc}
    \toprule
    \multirow{2}{*}{Benchmark} & \multirow{2}{*}{Software} & \multirow{2}{*}{Task} & \multirow{2}{*}{Platform} & \multirow{2}{*}{Env?} & \multirow{2}{*}{Inst. Video?} & \multirow{2}{*}{GT Plan} & Task State \\
     & & & & & & & variability?\\
    \midrule
    WebArena \citep{webarena} & 6 & 812 & Web & Yes & \xmark{} & \xmark{} & \xmark{} \\
    VisualWebArena \citep{visualwebarena} & 3 & 910 & Web & Yes & \xmark{} & \xmark{} & \xmark{}\\
    WebVoyager \citep{webvoyager} & 15 & 643 & Web & Yes & \xmark{} & \xmark{} & \xmark{} \\
    AutoDroid \citep{AutoDroid} & 13 & 158 & Android OS & Yes & \xmark{} & \xmark{} & \xmark{}\\
    AndroidWorld \citep{androidworld} & 20 & 116 & Android OS & Yes & \xmark{} & \xmark{} & \xmark{}\\
    AgentStudio \citep{zheng2025agentstudio} & 9 & 205 & Desktop + Web & Yes & \xmark{} & \xmark{} & \xmark{} \\
    Mobile-Eval \citep{wang2024mobile} & 10 & 30 & Android OS & Yes & \xmark{} & \xmark{} & \xmark{}\\
    APPAgent \citep{zhang2023appagent} & 10 & 50 & Android OS & Yes & \xmark{} & \xmark{} & \xmark{}\\
    OSWorld \citep{OSWorld} & 10 & 369 & Desktop & Yes & \xmark{} & \xmark{} & \xmark{} \\
    \midrule
    AssistGUI \citep{assistgui} & 9 & 100 & Windows & No & \cmark{} & \xmark{} & \xmark{} \\
    WindowAgentArena \citep{windowsagentarena} & 11 & 154 & Windows & Yes & \xmark{} & \xmark{} & \xmark{}\\
    \textbf{\bench{}} & 10 & 611 & Win. + Web & No & \cmark{} & \cmark{} & \cmark{} \\
    \bottomrule
    \end{tabular}
    }
    \vskip -0.2in
\end{table*}

\section{Introduction}

Graphical User Interface (GUI) automation has emerged as a promising direction for productivity-oriented AI systems. 
With the rapid advancement of Multimodal Large Language Models (MLLMs) such as GPT-5~\citep{gpt-5} and Claude-4-Sonnet~\citep{claude4}, GUI agents have demonstrated the potential to assist users in software usage, file management, office design, coding, and web browsing, reducing repetitive work and improving workflow efficiency.

Unlike traditional computer vision tasks such as image recognition~\citep{he2016deep} or visual question answering~\citep{antol2015vqa, VQAv2}, GUI automation operates in a highly \emph{dynamic} environment where the interface state evolves as a result of user actions, system configurations, or partial task progress. 
However, existing online GUI benchmarks, including WebArena~\citep{webarena}, WebVoyager~\citep{webvoyager}, OSWorld~\citep{OSWorld}, AssistGUI~\citep{assistgui}, and WindowsAgentArena~\citep{windowsagentarena}, primarily evaluate agents from a single, canonical initial state and measure success only at the end of the trajectory. 
Such \emph{static} evaluation overlooks several common situations in real-world usage:  
(1) software interfaces are often not in their default configurations,  
(2) users frequently request assistance from intermediate states of partially completed tasks, and  
(3) agents with the same low success rate (e.g., 20\%) may differ substantially in their robustness, planning stability, and self-correction behavior differences that cannot be revealed under a single start-state protocol.

In parallel, robustness-oriented GUI benchmarks such as GUI-Robust~\cite{yang2025guirobust} examine how agents respond to \emph{interface-level} perturbations, including layout shifts, appearance variations, and transient pop-ups. 
These benchmarks vary the visual presentation while keeping the task goal and initial state fixed, providing valuable insights into perception robustness. 
In contrast, \bench{} does not alter UI skin, resolution, or layout; instead, it targets \emph{task-progress–level} variation by modifying the executed steps themselves (Add-, Trim-, and Adjust-step). This form of task-state variability, such as missing, redundant, or reordered steps, poses challenges fundamentally different from interface perturbations, making the two robustness dimensions complementary and largely orthogonal.

To complement existing benchmarks, we focus on robustness to \emph{intermediate starting states}, a ubiquitous but under-explored aspect of GUI interactions.  
We introduce \bench{}, a benchmark that systematically constructs diverse initial states for each task by applying controlled \emph{pre-actions}.  
These pre-actions modify task progress itself (e.g., steps partially completed, over-executed, or executed in a different order), allowing tasks with the same goal to begin from multiple realistic contexts.  
This design mirrors how real users summon assistance in the middle of workflows and enables diagnostic evaluation of an agent’s ability to recover, adapt plans, and handle contextual variability.  
\bench{} covers 10 widely used desktop applications and provides 611 tasks, each paired with a user query, instructional video, and project file. 
To construct meaningful state variations, trained annotators first demonstrate ground-truth (GT) plans, after which we generate augmented initial states via pre-actions.

In addition, we propose \agent{}, a simple and general GUI agent framework grounded in \emph{critical-thinking} principles, an aspect that has received relatively limited attention in prior GUI agents~\citep{hong2024cogagent, cheng2024seeclick, autowebglm, agents1, osatlas}.  
Dynamic GUI environments often deviate from expected states due to user behavior or system configurations, making it essential for agents to detect mismatches and adapt their plans.  
Through an analysis of real-world GUI scenarios, we distill three design principles that we argue are fundamental for GUI automation:  
\textbf{(1) Post-Planning Critique},  
\textbf{(2) Pre-Action Validation}, and  
\textbf{(3) Post-Action Evaluation}.  
These components enable \agent{} to refine plans, verify step prerequisites, and evaluate action outcomes, thereby improving reliability in dynamic settings.

Experiments reveal three key observations:  
(1) a substantial gap remains between humans and current agents, with models such as UI-Tars and GPT-5.1 still struggling on our curated desktop GUI tasks;  
(2) performance degrades sharply under non-default initial states, highlighting limited robustness and fragile planning capabilities; and  
(3) \agent{} significantly improves robustness and overall task success across \bench{} and WindowsAgentArena, even when built upon weaker base models and without prompt tuning, demonstrating the importance of multi-level critique for dynamic GUI automation (see Section~\ref{sec:exp}).

%% file: sections/8_related.tex
\section{Related Work}

\begin{figure*}[t]
\centering
% \vskip -0.2in
\includegraphics[width=0.86\linewidth]{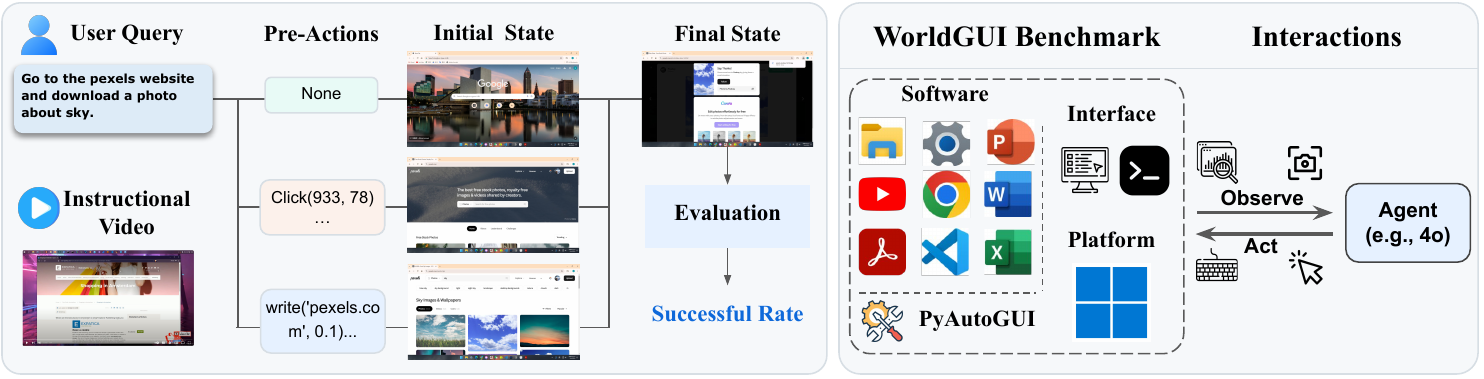}
\vskip -0.1in
\caption{\textbf{\bench{}.} Left: \bench{} creates pre-actions for each meta task, leading to different initial states. It successfully reflects the real-world human-computer interaction process. Right: components in \bench{}.}
\label{fig:benchoverview}
\vskip -0.2in
\end{figure*}

\subsection{GUI Benchmarks}

GUI benchmarks provide the foundation for evaluating computer-use agents across web, desktop, and mobile environments. 
Early interactive benchmarks such as WebShop~\citep{webshop}, WebArena~\citep{webarena}, and WebVoyager~\citep{webvoyager} center on browser tasks with long-horizon goals. 
OSWorld~\citep{OSWorld} and mobile benchmarks, including MobileAgent~\citep{wang2024mobile} and AppAgent~\citep{zhang2023appagent} broaden coverage to multi-application desktop and mobile ecosystems. 
AssistGUI~\citep{assistgui} and WindowsAgentArena \cite{windowsagentarena} provide online execution environments for Windows-based tasks.

Recent benchmarks emphasize scalability and online evaluation.  
UIExplorer \cite{UIExplorer} offers a large-scale environment for GUI exploration in both structured-DOM and pure screen modes.  
GUI-World \cite{guiworld2025} provides multi-application desktop tasks with online execution and standardized initial states.  
Robustness-oriented suites such as GUI-Robust \cite{yang2025guirobust} examine agents under interface-level perturbations (e.g., layout shifts, visual noise, pop-ups), focusing on perception and action robustness. While these benchmarks significantly advance GUI automation, they overwhelmingly assume a \emph{single canonical initial state} for each task. They do not explicitly evaluate \emph{task-state variability}, such as partially completed workflows, missing or redundant steps, or alternative intermediate contexts common in real user workflows. Our benchmark, \bench{}, complements existing online environments by systematically constructing diverse initial states for the same task goal via controlled pre-actions, enabling diagnostic evaluation of robustness, recoverability, and planning stability in dynamic GUI environments.

\subsection{GUI Agents}

Recent progress in GUI agents spans perception, grounding, planning, and full computer-use control. 
CogAgent~\citep{hong2024cogagent} improves GUI perception, while SeeClick~\citep{cheng2024seeclick} and SeeAct~\citep{zheng2024seeact} focus on grounding UI elements for more reliable action prediction. 
MobileAgent~\citep{wang2024mobile} and AppAgent~\citep{zhang2023appagent} extend these capabilities to mobile environments. On desktop and OS-level tasks, AssistGUI~\citep{assistgui} adopts a Plan-Act pipeline with a single post-action checker. 
Agent-S~\citep{agents1} introduces experience-augmented hierarchical planning with a Manager-Worker architecture, a self-evaluator, and an Agent-Computer Interface (ACI). 
Agent-S2~\citep{agents2} further incorporates specialist grounding (Mixture-of-Grounding) and Proactive Hierarchical Planning, achieving strong performance across OSWorld, WindowsAgentArena, and AndroidWorld. Beyond inference-time frameworks, several works train GUI agents or their components.  Aguvis~\citep{xu2024aguvis} and UGround~\citep{gou2024uground} train pure-vision or universal grounding models at scale.  
AgentTrek~\citep{xu2025agenttrek} synthesizes multi-step trajectories from web tutorials for supervised training, and SE-GUI~\citep{yuan2025segui} applies reinforcement learning to strengthen grounding robustness. While these approaches enhance perception, grounding, or hierarchical planning, they typically employ only a single verification stage (e.g., self-evaluation or post-action checking).  In contrast, our framework \agent{} focuses on \emph{structuring the verification process itself} for dynamic GUI environments. 
\agent{} introduces three complementary critique stages: \textbf{Post-Planning Critique}, \textbf{Pre-Action Validation}, and \textbf{Post-Action Evaluation}, enabling agents to detect state mismatches, assess step readiness, and repair execution failures. 
This multi-level critique design is orthogonal to hierarchical planners and training-based pipelines, and can be combined with them without additional model training.
A component-wise comparison is provided in Table~\ref{tab:compareagents}.

% \subsection{Critical Thinking in Agents}

% Recent advances in foundation models, including OpenAI-o1~\citep{openaio1} and DeepSeek-R1~\citep{deepseekr1}, highlight the effectiveness of structured reasoning and \emph{verify-then-correct} processes for challenging tasks.  
% LLM-based agent frameworks such as Reflexion~\citep{shinn2024reflexion} and CRITIC~\citep{gou2023critic} operationalize these ideas by analyzing intermediate outputs, detecting failures, and revising plans through iterative feedback loops. GUI automation naturally benefits from such mechanisms due to its long-horizon, stateful nature.  
% AssistGUI~\citep{assistgui} incorporates a post-action check to verify execution results; however, it does not extend critique to the planning or pre-execution stages.  
% Building upon the general philosophy of verify–then–correct, we introduce \agent{}, which integrates three complementary critique modules: 
% \textbf{(1) Post-Planning Critique},  
% \textbf{(2) Pre-Action Validation}, and  
% \textbf{(3) Post-Action Evaluation}.  
% Together, these components form a unified and broadly applicable framework for robust GUI control, enabling agents to refine plans, ensure step readiness, and detect execution anomalies in dynamic environments.

%% file: sections/3_bench.tex
\section{\bench{} Benchmark}
\label{sec:benchmark}

Following prior work~\cite{OSWorld, webarena}, we model GUI automation as a partially observable Markov decision process (POMDP).  
A complete specification of the POMDP components and the execution environment is provided in Appendix~\ref{sec:appendix_task_details}.

\subsection{Data Source}

\bench{} consists of a broad spectrum of widely-used desktop applications, which can be categorized into five main groups: \textit{(i)} Office, includes PowerPoint, Word, Excel, and Adobe Acrobat; \textit{(ii)} Windows Usage, includes System Settings and File Management; \textit{(iii)} Web Usage, includes the configuration of Youtube and website operations; \textit{(iv)} Coding, focus on the customization, configuration and editing of Visual Studio Code (VSCode); \textit{(v)} Media, operating VLC player for video editing and creation.

\begin{table*}[t]
    \vskip -0.1in
    \caption{Task category, task activities, and project file of the desktop applications in \bench{}.}
    \label{tab:dataoverview}
    \centering
    \vskip -0.1in
    \resizebox{0.86\linewidth}{!}{
    \begin{tabular}{llllll}
    \toprule
      Category & Applications  & All Task & Avg. Length & Task Activities & Project File Type \\
    \midrule
       Office & PowerPoint & 64 & 7.1 & Change the content style and layout; Design new effects & project.pptx\\
        Office & Word & 63 & 4.1 & Formatting the content style and layout & project.docx \\
        Office & Excel & 70 & 5.1 & Table formatting; Data management and processing & project.xlsx\\
        Office & Adobe Acrobat & 66 & 5.1 & Automatic add electric signature; Document management & project.pdf\\
        Coding & VSCode & 56 & 4.4 & Code editing; Software configuration & vscode.exe\\
        Windows Usage & Settings & 69 & 7.1 & Advanced personalized and safety settings; & ms-Settings\\
        Windows Usage & File Explorer & 44 & 5.1 & File management: Add, delete, rename, and move files & explorer.exe\\
        Web Usage & Web Browser & 59 & 7.8 & Web operation & web browser + URL\\
        Web Usage & Youtube (Online) & 61 & 4.8 & Video and account configurations & web browser + URL\\
       Media & VLC Player & 59 & 10.8 & Video editing and creation & project.mp4\\
       \hline
       Total (Average) & -- & 611 & 6.0 &  & \\
    \bottomrule
    \end{tabular}
    }
    \vskip -0.1in
\end{table*}

\begin{figure*}[t]
    \centering
    \includegraphics[width=0.3\linewidth]{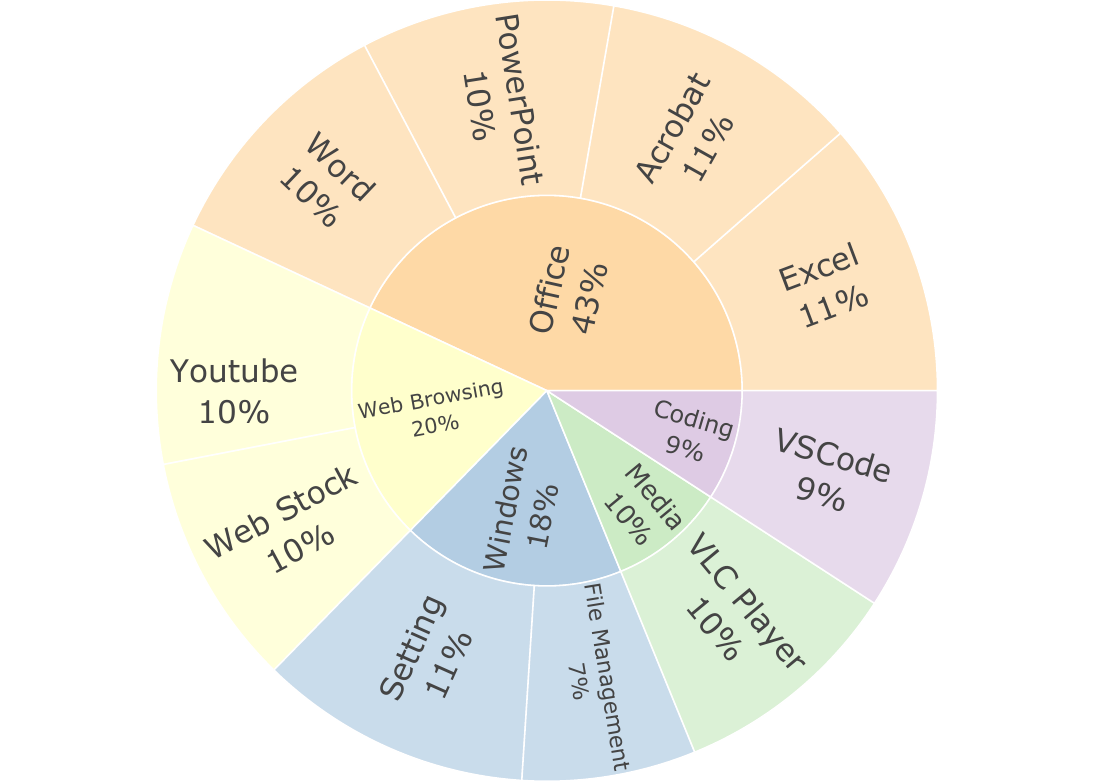}
    % \hfill
    \includegraphics[width=0.38\linewidth]{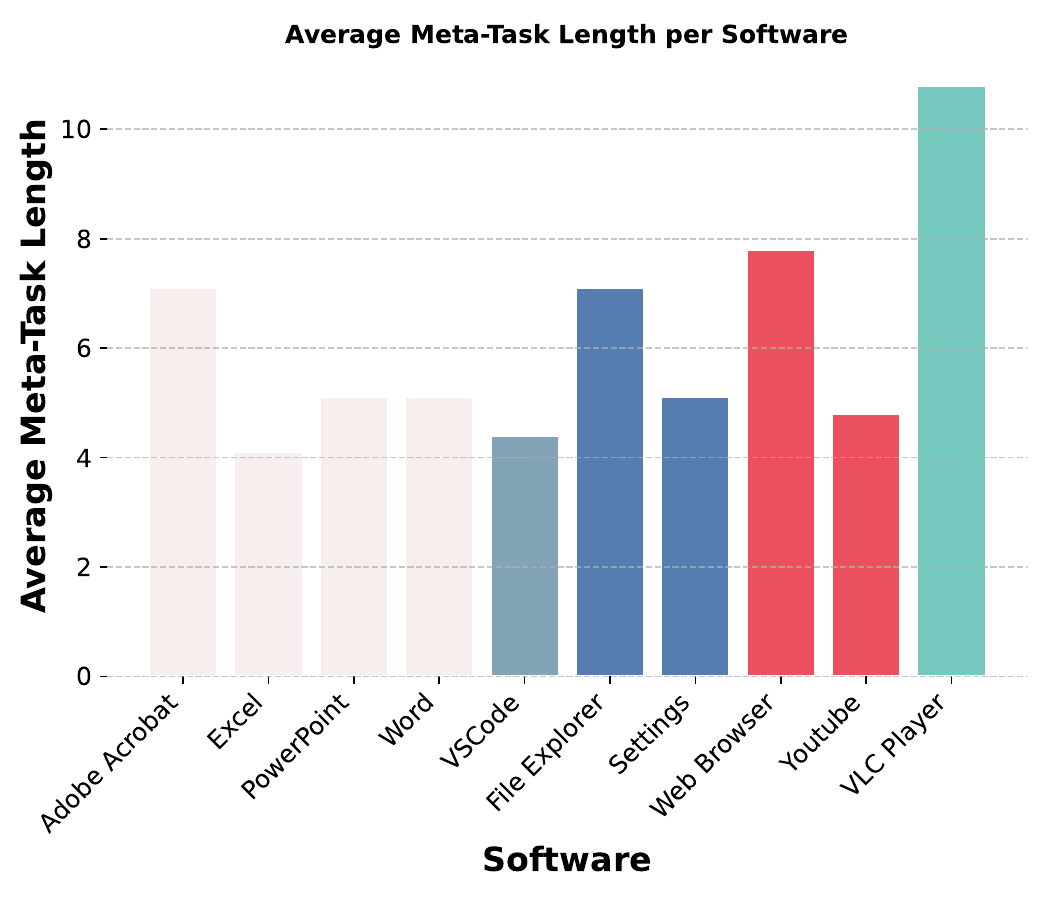}
    % \hfill
    \caption{The left shows software taxonomy in \bench{}. The right shows the distribution of task length.}
    \label{fig:datastats}
    \vskip -0.2in
\end{figure*}

\subsection{Pipeline of Data Construction} 

We engaged four annotators and developed the necessary scripts to structure and format the data. To reduce annotator-style variance, all annotators followed a standardized annotation protocol. Additionally, to facilitate ground truth (GT) plan generation and pre-action generation, we implemented simple agent-based methods to collect the relevant data. The overall data construction pipeline comprises six steps, as detailed below.

\noindent\textbf{Raw Video Collection.} We collect raw videos from the YouTube website as there are a lot of high-quality tutorials for desktop applications with high views. For each software, we ask the annotators to watch the videos first and download them via the diversity of software usage.

\noindent\textbf{Instruction Video Preparation.} After obtaining the raw videos, we write the script codes to cut the lengthy and noisy videos into the sub-clips (30 seconds to 3 minutes) that serve as the instructional video.

\noindent\textbf{User Query Generation.} After obtaining the instructional videos, annotators are asked to manually write user queries corresponding to each video. For example, a user query for a task involving File Explorer might be: \textit{``Please compress the project.mp4 into an MPEG-4 file optimized in 1080p.''}

\noindent\textbf{Project File Preparation.} Following the AssistGUI~\citep{assistgui}, we create the project file for each task to ensure reproducibility without relying on resource-intensive virtual machines~\citep{OSWorld} or Docker environments~\citep{windowsagentarena}. This approach guarantees that the testing process begins from a consistent state. When combined with pre-actions, it enables augmentation of the same task with various initial states.

% \noindent\textbf{GT Plan Generation.} We write the script to accept user query $q$ and instructional video $v$ as input and generate the raw plans by the agent (powered by GPT-4o). Since the raw plans are not flawless, annotators are asked to watch the videos and manually execute the tasks following the raw plans. During this process, annotators edit the plans to correct any inaccurate steps or descriptions, ultimately producing the finalized GT plans. \textbf{It is noted that GT plans are only used in creating pre-actions. We do not use them in running agents.}

\noindent\textbf{GT Plan Generation.} Given a user query $q$ and instructional video $v$, we automatically generate an initial draft plan using a GPT-4o–based agent. 
Annotators then watch the video and execute the task while reviewing the draft plan, correcting any inaccurate or missing steps to obtain the final GT plan. 
\textbf{GT plans are used solely for constructing pre-actions and are never provided to or used by any evaluated agent.}

\begin{figure}[t]
\centering
\includegraphics[width=0.9\linewidth]{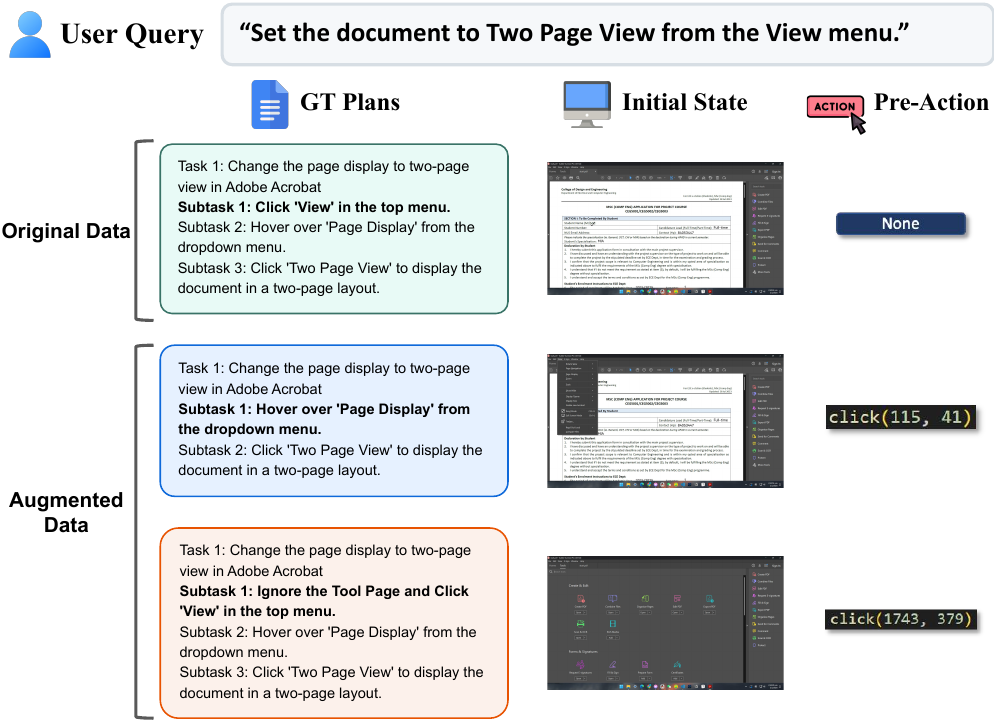}
\caption{An example of augmenting one GUI task with manually aug the initial state and then using the execution scripts and corresponding agents to obtain the pre-action for each augmented case.}
\label{fig:dataaugment}
\vskip -0.2in
\end{figure}

\noindent\textbf{Pre-Actions Generation.} To vary the task, we propose introducing pre-actions before the task begins. These pre-actions are created by annotators and involve corresponding scripts and agents. They are written in Python code, for example:
\texttt{from pyautogui import click, rightClick\textbackslash n rightClick(800,400)}. 
The pre-actions primarily serve two purposes: \textbf{1) Simulating Intermediate Task States}: Pre-actions can complete specific steps of a task, creating a starting point from an intermediate state. This approach addresses scenarios where users may invoke GUI assistant at any time. For example, if the task involves opening a dropdown menu, the pre-action may pre-open the menu. If the agent fails to recognize this precondition and follows its plan to click the menu again, it might inadvertently close the menu, causing task failure.
\textbf{2) Introducing Diverse Initial Context States}: Pre-actions can also introduce variations in the initial state, such as opening random tabs or settings. This ensures that the starting state is unconventional, challenging the agent to adapt by modifying its plan or adding necessary new steps. See example in Figure~\ref{fig:dataaugment}.
Each pre-action was manually verified to ensure the task remained solvable.

\subsection{Data Statistics}

\textbf{Overall.} We present the detailed task taxonomy and lengths about \bench{} in Figure \ref{fig:datastats} and Table \ref{tab:dataoverview}. A total of 111 meta tasks were collected from these applications, with each task being augmented ~5 times based on the task's functionality, resulting in 500 augmented tasks. In total, \bench{} comprises 611 tasks, and every task has almost 6 variation instances, which is capable of reflecting the real-world interactions of the GUI environment. \textbf{All data are available for reference.}\footnote{\url{https://huggingface.co/datasets/anonymousABC/WorldGUI-Bench}}

\begin{figure}[t]
    \centering
    \includegraphics[width=0.9\linewidth]{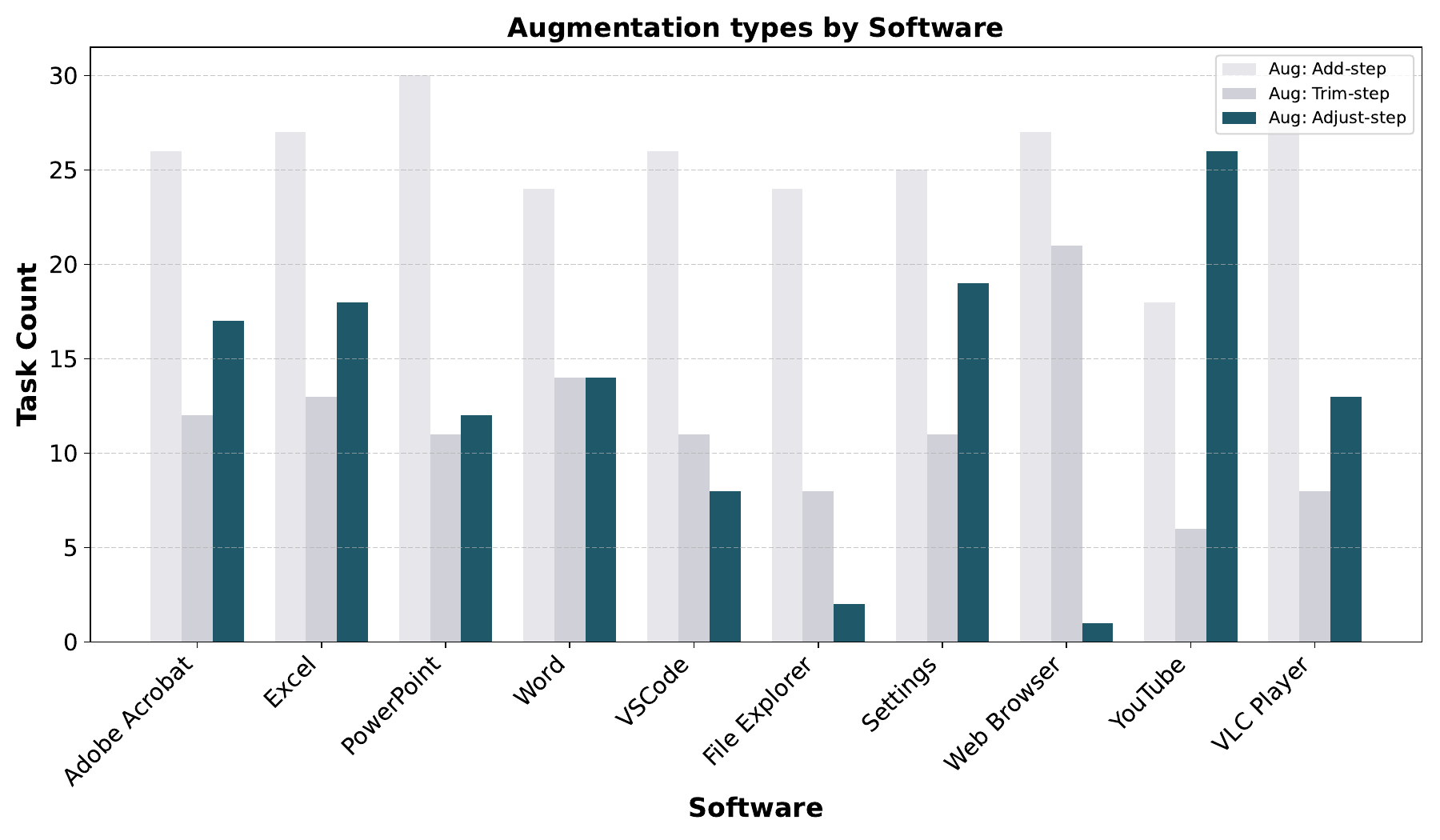}
    \vskip -0.1in
    \caption{The distribution of different augmentation types.}
    \label{fig:augtype}
    \vskip -0.1in
\end{figure}

\begin{figure}[t]
    \centering
    \includegraphics[width=0.9\linewidth]{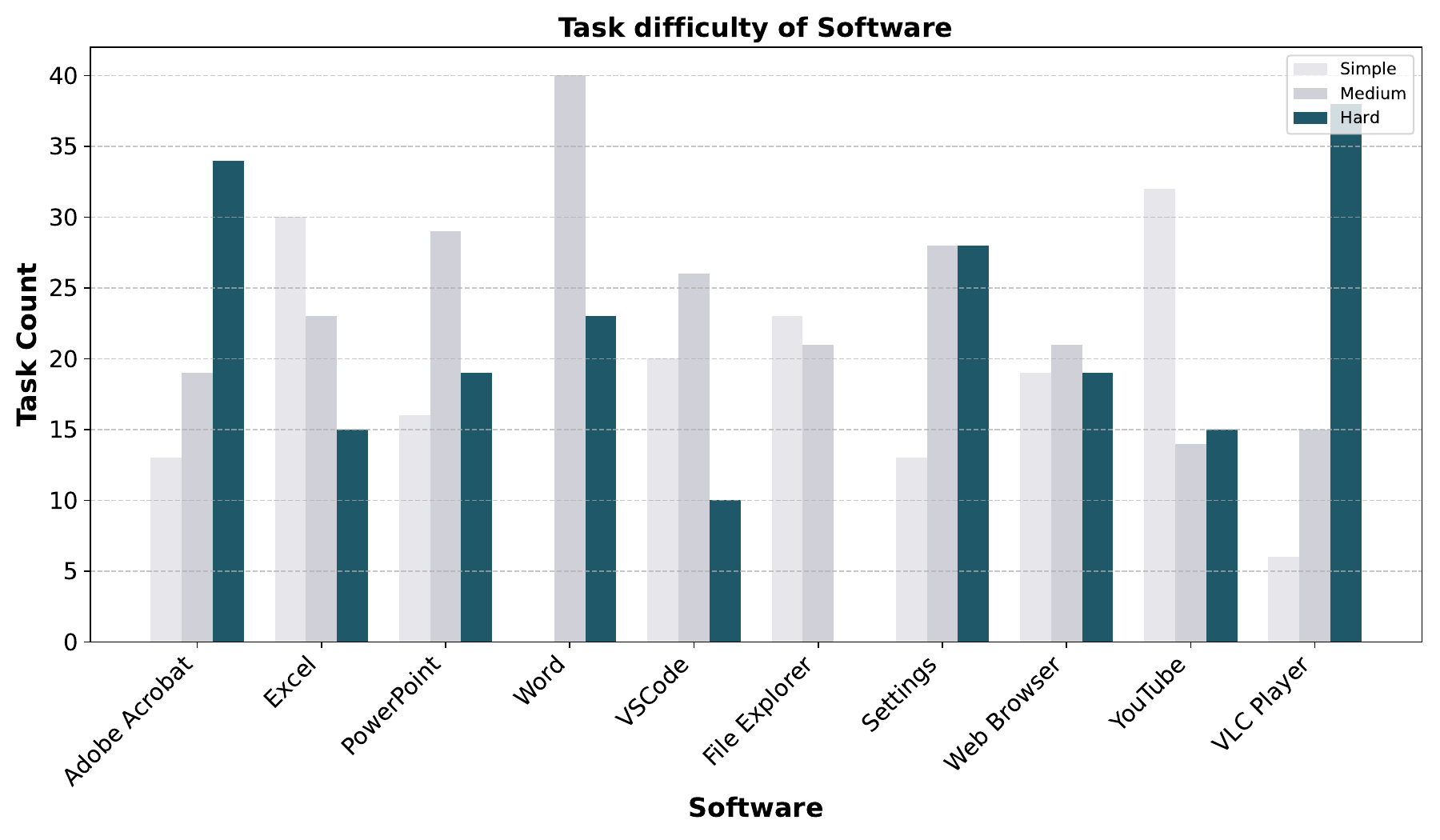}
    \vskip -0.1in
    \caption{The distribution of different task difficulty.}
    \label{fig:taskdiff}
    \vskip -0.1in
\end{figure}

\textbf{Augmentation tasks type analysis.} As we summarize, the real GUI scenarios include: (1) The software interface is not in its default state. (2) The human-computer interactions may start from the intermediate state of a specific task. Our augmentations lie in two main groups: (1) simulating the intermediate states and (2) introducing diverse initial states. We divide the two groups into three types: \textbf{Add-step}, \textbf{Trim-step}, and \textbf{Adjust-step}. For Add-step, it represents various unrelated state augmentations to simulate the scenario that we may start the task in another unrelated task or interfaces, the agent should replan the task to add necessary steps. For Trim-step, it represents that we finish several steps of a long task and make the task in an intermediate state. For Adjust-step, it is usually a small modification of the existing state, such as changing the interface by clicking another Tab or clicking a button to open an unrelated dropdown menu. Most of the time, it would not require new steps to return to the target task procedure. This augmentation may mislead the agent in state understanding, making them jump or miss the key steps.  As shown in Figure \ref{fig:augtype}, the manually created augmentations mainly belong to the add-step. Adjust-step could be the second-largest part. Due to the low complexity of the interfaces of File Explorer, we cannot create many augmentations for adjust-step. 

\textbf{Task difficulty analysis.} Figure \ref{fig:taskdiff} shows the distribution of the task difficulty across desktop applications. We annotated the task difficulty level based the subjective software usage experience. The results indicate that the tasks in Adobe Acrobat and VLC player are more challenging. The tasks in Excel, PowerPoint, and Word are more at the medium and simple levels. By considering the Success Rate and task length on these tasks, one can know that the tasks are easy for humans, but hard for current GUI agents. Overall, the task difficulty of our created data is diverse across different applications.

%% file: sections/5_exp.tex
\section{WorldGUI-Agent}

\begin{table}[h]
  \centering
  \vskip -0.1in
  \caption{Comparison with closely related GUI agents.}
  \vskip -0.1in
  \label{tab:compareagents}
  \scalebox{0.8}{
  \begin{tabular}{lccc}
    \toprule
    Method & Post-Plan & Pre-Action & Post-Action \\
    \midrule
    Mobile-Agent / V2     & \xmark{} & \xmark{} & \cmark{} \\
    AssistGUI             & \xmark{} & \xmark{} & \cmark{} \\
    Agent-S / S2          & \xmark{} & \xmark{} & \cmark{} \\
    Mobile-Agent-E & \xmark{} & \xmark{} & \cmark{} \\
    \agent{} (ours)       & \cmark{} & \cmark{} & \cmark{} \\
    \bottomrule
  \end{tabular}
  }
  \vskip -0.1in
\end{table}

To study robustness under the dynamically perturbed initial states of \bench{}, we adopt a lightweight, non-training agent framework. As summarized in Table~\ref{tab:compareagents}, most existing agents rely on a single post-action evaluation module. Rather than proposing a new algorithm, our aim is simply to structure the standard perception–planning–execution loop into clearer verification stages.\agent{} introduces three small modules at different abstraction levels:  
\textbf{Planner-Critic} refines the draft plan;  
\textbf{Step-Check} uses the screenshot and metadata to decide whether a plan step should be executed, skipped, or rewritten;  
\textbf{Actor-Critic} validates each action using the current observation and recent action history. Compared to prior self-reflection approaches (e.g., Reflexion, CRITIC), which apply a single global critique, our checks are lightweight and distributed across plan, step, and action levels to better handle task-progress variation. Full module specifications and I/O formats are included in Appendix~\ref{sec:agent}.

\section{Experimental Results}
\label{sec:exp}

\begin{table*}[!t]
    \renewcommand{\arraystretch}{1.1}
    \centering
    % \vskip -0.2in
    \caption{Success rate (\%) of different agents on \bench{}. Human$^*$ denotes the average performance of four expert participants who have watched the instructional video only once, similar to the model. Meta represents the meta task, while Aug. represents the augmented task.}
    \vskip -0.1in
    \scalebox{0.72}{
    \begin{tabular}{l|cc|cc|cc|cc|cc|c}
    \toprule
    \multirow{2}{*}{\textbf{Method}} & \multicolumn{2}{c}{\textbf{Office}} & \multicolumn{2}{c}{\textbf{Win. Usage}} & \multicolumn{2}{c}{\textbf{Web}} & \multicolumn{2}{c}{\textbf{Coding}} & \multicolumn{2}{c|}{\textbf{Media}} & \multirow{2}{*}{\textbf{Overall}}\\
    & Meta & Aug. & Meta & Aug. & Meta & Aug. & Meta & Aug. & Meta & Aug. & \\
    \midrule
    Plan-Act w/ Gemini-2.0 & 8.9 & 3.2 & 8.3 & 3.4 & 28.6 & 16.2 & 18.2 & 2.2 & 10.0 & 2.0 & 6.9 \\
    Plan-Act w/ GPT-4o & 13.3 & 10.1 & 8.3 & 2.3 & 23.8 & 11.1 & 9.1 & 2.2 & 10.0 & 2.0 & 8.5\\
    AssistGUI w/ GPT-4o & 26.7 & 16.1 & 29.2 & 7.9 & 33.3 & 20.2 & 27.3 & 11.1 & 10.0 & 8.2 & 16.5\\
    CCU w/ Claude-3.5-Sonnet & 28.9 & 19.3 & 29.2 & 14.6 & 71.4 & 32.3 & 54.6 & 22.2 & 30.0 & 6.1 & 23.6\\
    UI-TARS-1.5 & 28.9 & 16.1 & 12.5 & 2.2 & 28.6 & 9.1 & 36.7 & 6.7 & 0.0 & 0.0 & 12.3\\
    Agent S2 & 33.3 & 16.5 & \textbf{70.8} & \textbf{59.6} & 52.4 & 45.5 & 45.5 & 37.8 & 20.0 & 16.3 & 34.2\\
    \midrule
    \multicolumn{12}{c}{\textbf{\agent{} (Ours)}} \\
     w / Gemini-2.0 & 31.1 & 17.0 & 20.8 & 9.0 & 38.1 & 29.3 & 36.4 & 11.1 & 20.0 & 10.2 & 19.1 \\
     w / GPT-4o & 42.2 & 24.3 & 41.7 & 11.2 & 47.6 & 35.4 & 45.5 & 15.6 & 40.0 & 12.2 & 26.0 \\
     w / Claude-3.5-Sonnet & 57.8 & 32.6 & 50.0 & 19.1 & 76.2 & 46.5 & 54.6 & 26.7 & 50.0 & 18.4 & 36.0 \\
     w / GPT-5.1 & \textbf{64.4} & \textbf{37.6} & 62.5 & 40.4 & \textbf{85.7} & \textbf{53.5} & \textbf{63.6} & \textbf{46.6} & \textbf{60.0} & \textbf{26.5} & \textbf{45.8}\\
    \midrule
    \textbf{Human$^*$} & 88.9 & 83.5 & 100.0 & 89.9 & 95.2 & 80.8 & 81.8 & 77.8& 90.0 & 85.7 & 85.3\\
    \bottomrule
    \end{tabular}
    }
    % \vskip -0.2in
    \label{tab:mainresult}
\end{table*}

\begin{table*}[ht]
    \centering
    % \vskip -0.1in
    \caption{Success rate (\%) of our \agent{} (w/ GPT-4o) with the ablation of different critical modules.}
    \vskip -0.1in
    \scalebox{0.72}{
    \begin{tabular}{l|cc|cc|cc|cc|cc|c}
    \toprule
    \multirow{2}{*}{\textbf{Method}} & \multicolumn{2}{c}{\textbf{Office}} & \multicolumn{2}{c}{\textbf{Win. Usage}} & \multicolumn{2}{c}{\textbf{Web}} & \multicolumn{2}{c}{\textbf{Coding}} & \multicolumn{2}{c|}{\textbf{Media}} & \multirow{2}{*}{\textbf{Overall}}\\
     & Meta & Aug. & Meta & Aug. & Meta & Aug. & Meta & Aug. & Meta & Aug. & \\
    \midrule
    \rowcolor{graybg} Full Model & 42.2 & 24.3 & 41.7 & 11.2 & 47.6 & 35.4 & 45.5 & 15.6 & 40.0 & 12.2 & 26.0 \\
    -- w/o Planner-Critic & 31.1 & 17.0 & 20.8 & 9.0 & 38.1 & 25.3 & 36.4 & 11.1 & 20.0 & 10.2 & 18.5\\
    -- w/o Step-Check & 31.1 & 19.3 & 20.8  & 9.0 & 33.3 & 28.3 & 45.5 & 13.3 & 20.0 & 8.2 & 19.8 \\
    -- w/o Actor-Critic & 15.6 & 7.8 & 4.2 & 3.4 & 28.6 & 17.2 & 0.0 & 8.9 & 10.0 & 6.1 & 9.7 \\
    \bottomrule
    \end{tabular}
    }
    % \vskip -0.1in
    \label{tab:ablationmodule}
\end{table*}

\begin{table}[ht]
    \vskip -0.1in
    \centering
    \caption{Success rate (\%) of our \agent{} with the ablation of Instructional Video.}
    \vskip -0.1in
    \scalebox{0.64}{
    \begin{tabular}{l|ccccc|c}
    \toprule
     \textbf{Method} & \textbf{PPT}& \textbf{Word} & \textbf{Excel} & \textbf{Acrobat} & \textbf{VSCode} & \textbf{Overall}\\
    \midrule
    \rowcolor{graybg} Full Model & 45.5 & 36.4 & 50.0 & 36.4 & 45.5 & 42.9\\
    w/o Inst. Video & 45.5 & 27.3 & 25.3 & 18.2 & 27.3 & 28.6\\
    \bottomrule
    \end{tabular}
    }
    \label{tab:noinstvideo}
\end{table}

\begin{table}[ht]
    \vskip -0.1in
    \centering
    \caption{Running time comparison.}
    \vskip -0.1in
    \scalebox{0.8}{
    \begin{tabular}{lll}
    \toprule
     \textbf{Method} & \textbf{Executed Steps} & \textbf{Time (seconds)} \\
    \midrule
    Agent-S & 10 & 131.98\\
    Agent-S2 & 9 & 108.65\\
    \agent{} & 24 & 129.55 \\
    \bottomrule
    \end{tabular}
    }
    \vskip -0.1in
    \label{tab:computation}
\end{table}

\begin{table}[t]
\centering
\vskip -0.1in
\caption{Task counts and success rates (SR, \%) across different augmentation types for \agent{} (GPT-5.1).}
\label{tab:aug_type_results}
\vskip -0.1in
\scalebox{0.74}{\begin{tabular}{lccc}
\toprule
\textbf{Augmentation Type} & \textbf{\#Tasks} & \textbf{SR (\%)} & \textbf{\#Success} \\
\midrule
Add-step     & 255 & 34.1 & 87 \\
Trim-step    & 119 & 52.1 & 62 \\
Adjust-step  & 130 & 43.1 & 56 \\
\midrule
\textbf{Total} & 500 & 41.0 & 205 \\
\bottomrule
\end{tabular}}
\vskip -0.1in
\end{table}

\textbf{Implementation Details.} We use PyAutoGUI\footnote{\url{https://pyautogui.readthedocs.io}} to extract basic metadata from GUI screenshots, including bounding boxes of buttons, icons, and text regions. 
We then implement software-specific GUI parsers\footnote{\url{https://anonymous.4open.science/r/WorldGUI-7A5C/agent/gui_parser/}} that further refine element grounding by incorporating OCR results obtained through the Google OCR API. 
Details of our baseline agent \agent{} are provided in Appendix~\ref{sec:agent}. 
All experiments are conducted under a fixed screenshot resolution of 1920$\times$1080 and a display scale of 125\%. 
For \agent{}, we limit the Actor–Critic correction attempts to three steps to control interaction cost. 
The total number of trials allowed per task is set to $4N + 1$, where $N$ is chosen empirically based on the task category.

\textbf{Evaluation.} Given that our \bench{} includes 611 GUI tasks, we engaged four participants with strong coding and software backgrounds to test all tasks and document their evaluation results. \textbf{Metric.} Following the previous works, we use Success Rate (SR) as the metric.

\textbf{Baselines.} We implement the baseline approach called \textbf{Plan-Act} with different MLLMs as the base model. It focuses on investigating the basic capabilities of task planning and action prediction. Additionally, we compare our \agent{} with two agentic frameworks and two SOTA GUI models: \textbf{AssistGUI} \citep{assistgui}, \textbf{Agent-S2} \citep{agents2}, \textbf{Computer Use (Claud-3.5-Sonnet)} \citep{claude3.5}, and \textbf{UI-TARS-1.5} \citep{uitars}. AssistGUI and Agent-S2 are two prominent agentic frameworks designed for Desktop GUI Automation, which can plan the task and then execute the task step by step by following the query. We increase the base model to GPT-4o for AssistGUI and Claude-Sonnet-4 of Agent-S2 for fair performance. \textbf{Claude Computer Use (CCU)} is the leading proprietary model specially designed for computer use. We use the open-source implementation OOTB \citep{hu2024dawnguiagentpreliminary} as the codebase and then add the subtitle of instructional videos into the input prompt for a fair comparison. We also implement our \textbf{\agent{}} with four different MLLMs to illustrate the effectiveness of our proposed universal agent framework.

\subsection{Main Results on \bench{}}
\textbf{Overall.} Table \ref{tab:mainresult} reports the success rates (SR) of different agents and human experts on our \bench{} benchmark, broken down by task type (Meta vs. Aug.) across five categories: Office, Win. Usage, Web, Coding, and Media. From these results we draw the following main conclusions.

\textbf{A large gap remains between agents and humans.} The best-performing agent (\agent{} with GPT-5.1) achieves an overall SR of only 45.8\%, which is less than half of the 85.3\% attained by human experts. This stark contrast underscores the difficulty of our tasks and the need for further advances in desktop GUI automation.

\textbf{Agents generalize poorly to augmented tasks.} 
Across all methods, performance on Augmentation tasks (which introduce interface or context variations) is substantially lower than on their corresponding Meta tasks. For example, Claude-3.5-Sonnet in the Win. Usage category attains 50.0\% on Meta tasks but drops to just 19.1\% on Aug. tasks. This highlights the importance to evaluate the agents on various initial conditions.

\textbf{Desktop GUI tasks pose a greater challenge than web tasks.} Every agent scores higher on Web tasks than on desktop application tasks. \agent{} with GPT-5.1, for instance, jumps from 85.7\% on Web Meta to only 62.5\% on Win. Usage Meta, and the gap widens on their Augmentation counterparts. Thus, desktop GUI tasks remain challenges than widely-studied web GUI tasks \cite{cheng2024seeclick, webvoyager}.

\textbf{\agent{} consistently outperforms a naive Plan-Act counterpart.} By incorporating our three critical modules into the planning and execution loop, \agent{} substantially improves success rates over the basic Plan-Act counterpart. Relative to Plan-Act, \agent{} raises overall SR by +12.2\% with Gemini-2.0, +17.5\% with GPT-4o, and +12.4\% with Claude-3.5-Sonnet, demonstrating the effectiveness of our critical modules proposed for \agent{}.

\textbf{Specialized GUI models transfer poorly to \bench{}.}
Despite achieving strong results on several existing GUI benchmarks, specialized models do not perform well in our setting. 
For example, UI-TARS-1.5 attains only 12.3\% overall SR on \bench{}, which is notably lower than general-purpose MLLMs when used within our framework (e.g., 36.0\% for \agent{} with Claude-3.5-Sonnet and 45.8\% with GPT-5.1), and even below the CCU baseline built on Claude-3.5-Sonnet (23.6\%).
Similarly, Agent-S2 reaches 34.2\% overall SR, still far from human performance and trailing behind \agent{} with GPT-5.1.
These results indicate that training pipelines and architectures tuned for single-start or less diverse benchmarks do not automatically transfer to the multi-start, dynamically perturbed settings of \bench{}. 
In our evaluation, specialized GUI models exhibit strong grounding but substantially weaker planning, which leads to low success rates even when provided with additional signals such as instructional-video subtitles.

\begin{table*}[t]
    \centering
    \renewcommand{\arraystretch}{1.2}
    % \vskip -0.1in
    \caption{Experimental results on WindowsAgentArena. The reported results are from the \citep{windowsagentarena} and \citep{agents2}.}
    \vskip -0.1in
    \resizebox{0.76\linewidth}{!}{
    \begin{tabular}{lccccccc}
    \toprule
     \textbf{Method} & \textbf{Office} & \textbf{Web} & \textbf{Win. System} & \textbf{Coding} & \textbf{Media} & \textbf{Win. Utils} & \textbf{Overall}\\
    \midrule
     Phi3-V \citep{windowsagentarena} & 0.0 & 6.9 & 8.3 & 0.0 & 6.2 & 0.0 & 3.5\\
     GPT-4o-mini \citep{windowsagentarena} & 0.0 & 14.9 & 8.3 & 0.0 & 0.0 & 0.0 & 4.2\\
     GPT-4o \citep{windowsagentarena} & 0.0 & 13.7 & 29.2 & 0.0 & 10.3 & 0.0 & 8.6 \\
     NAVI \citep{windowsagentarena} & 0.0 & 27.3 & 33.3 & 27.3 & \textbf{30.3} & 8.3 & 19.5 \\
     Agent S \citep{agents1} w/ GPT-4o & 0.0 & 13.3 & 45.8 & 29.2 & 19.1 & 22.2 & 18.2 \\
     Agent S2 \citep{agents2} w/ Claude-3.7-Sonnet & 7.0 & 16.4 & \textbf{}54.2 & \textbf{62.5} & 28.6 & 33.3 & 29.8 \\
    \hline
      \rowcolor{graybg} \textbf{\agent{} w/ Claude-3.5-Sonnet} & \textbf{7.0} & \textbf{53.3} & \textbf{45.8} & 33.3 & 28.6 & \textbf{33.3} & \textbf{31.2}\\
    \bottomrule
    \end{tabular}
    }
    \vskip -0.2in
    \label{tab:windowsagentarena}
\end{table*}

\subsection{Ablation Study}

\textbf{Impact of different critical modules.} Table \ref{tab:ablationmodule} reports an ablation study of \agent{} across five application categories. The full model achieves an overall success rate (SR) of 26.0\%. Removing any core component leads to notable degradation. Excluding the Planner-Critic reduces SR to 18.5\% (–7.5\%), highlighting its role in plan refinement. Removing Step-Check lowers SR to 19.8\% (–6.2\%), mainly affecting multi-step interaction scenarios, suggesting its effectiveness in intermediate error correction. The most severe impact comes from removing the Actor-Critic, which collapses SR to 9.7\% (–16.3\%), nearly eliminating performance on Coding and Win. Usage tasks, underscoring the necessity of action-level reward feedback. Overall, the three modules provide complementary benefits, enabling robust agent performance.

\textbf{Impact of Instructional Video.} In Table \ref{tab:noinstvideo}, we study the impact of removing the instructional video (subtitles) by modifying the prompt to include only the user query for generating the initial plan. In the Excel applications, we observe a significant performance decline, as their tasks are complex and difficult, and rely more heavily on additional domain knowledge for successful planning. In contrast, the agents performs relatively well on Win. Usage tasks, such as Settings and File Management, are where it has more inherent pretrained prior. These findings underscore the necessity of instructional videos for complex tasks like visual effect design, mirroring how users learning to build a slide often rely on tutorial videos. It is note that it is a setup proposed in our benchmark; one can also use our data without instructional videos. 

\subsection{Results by Augmentation Type}

Across the 500 augmented tasks, \agent{} (GPT-5.1) solves 205 cases. 
As shown in Table~\ref{tab:aug_type_results}, Add-step tasks are the most challenging (34.1\% SR), since inserted steps create larger state mismatches. 
Adjust-step tasks show moderate difficulty (43.1\%), reflecting the need for fine-grained manipulations. 
Trim-step tasks are relatively easier (52.1\%), as removing redundant steps keeps the interface closer to the meta configuration. 
These trends highlight how different forms of task-progress variation distinctively impact agent robustness.

\subsection{Failure Analysis and Robustness Dimensions}

To better understand the limitations of current GUI agents under the dynamic initial states of \bench{}, we analyze representative failure modes drawn from our qualitative study (Appendix~\ref{sec:qualitative_results}). 
We observe three major robustness dimensions that frequently lead to failures.

\textbf{(1) State-mismatch robustness.} 
When pre-actions introduce interface configurations different from the default view (e.g., an expanded settings dropdown or partially filled input fields), agents often misinterpret the visible state or fail to recognize that previous steps have already been executed. 
As illustrated in Figure~\ref{fig:failurecases1} (left), once a submenu overlays other elements, the agent struggles to recover the intended target (e.g., the “System” button), leading to incorrect plan continuation or premature termination.

\textbf{(2) Fine-grained manipulation robustness.}
Tasks requiring precise adjustments—such as dragging sliders or manipulating bars to reach a specific value—remain challenging. 
Even when grounding is correct, the agent may overshoot, undershoot, or repeatedly click without achieving the required adjustment, as shown in Figure~\ref{fig:failurecases1} (right). 
This reflects a broader difficulty in performing continuous or multi-step manipulations that depend on spatial granularity.

\textbf{(3) Ambiguous-visual-context robustness.}
In the absence of clear text cues, agents exhibit uncertainty in selecting visually similar elements. 
Figure~\ref{fig:failurecases2} (left) illustrates a case where the model must choose a centered icon from a symmetric layout, yet struggles due to limited visual discrimination without textual anchors. 
Similarly, when multiple text elements appear close together, GUI parsers may surface misleading anchors (e.g., the label “Replace with”), causing the agent to click an irrelevant surface-level text region rather than the actual input box (Figure~\ref{fig:failurecases2}, right).

Overall, these failure modes echo our quantitative findings: while grounding on static interfaces is often strong, robustness to state changes, fine-grained manipulations, and text-sparse visual contexts remains limited.  
These observations highlight the need for future progress in both perception and planning components, particularly under the multi-start, dynamically perturbed conditions introduced by \bench{}.

\subsection{Computational Costs Analysis}

We select two SoTA non-training agent frameworks Agent-S \cite{agents1} and Agent-S2 \cite{agents2} for comparing the computational costs. Take a Windows setting task as an example, we tested on the same desktop PC with an AMD Ryzen 7 5800H CPU. As shown in Table \ref{tab:computation}, our \textbf{WorldGUI-Agent} shows a competitive running time of 129.55s, as compared with Agent-S (131.98s) and Agent-S2 (108.64s). The main computational costs of our designed modules are largely affected by calling base model. Since the main problem of desktop GUI automation is still the suboptimal performance as shown in Table \ref{tab:mainresult}, such computational costs are currently acceptable. There remains a clear tradeoff between performance and time costs in GUI automation, and this challenge is shared across the community.

\subsection{Results on WindowsAgentArena}

\textbf{Settings.} The official WindowsAgentArena (WAA) evaluation runs inside a Linux-based VM using a pre-built image and a JSON configuration file. Since the essential logic of WAA lies in its JSON specification of the environment state and task instruction, we reconstruct the same environment on a Windows machine by following the provided configuration (e.g., opening specified tabs or launching required applications).
We then use the instruction field from the JSON file as the prompt for our \agent{}, which executes the corresponding GUI actions on Windows. After completion, we follow WAA’s evaluation protocol to determine task success. When an official evaluation script is available, we apply it directly; otherwise, we perform manual judgment based on the given instructions. We do not optimize the prompts, just apply our agent directly on its tasks. \textbf{Main results.} Table \ref{tab:windowsagentarena} compares \agent{} against six agents on the WAA benchmark. \agent{} achieves a 31.2\% overall SR surpass two SOTA api-calling agents, Agent-S and Agent-S2, even with a weaker base mode,l Claude-3.5-Sonnet. These results underscore the effectiveness of our designed critical modules.

%% file: sections/6_con.tex
% \section*{Conclusion} In this paper, we take a first step toward comprehensive evaluation of GUI agents by introducing \bench{}. Beyond conventional static testing, we incorporate dynamic testing procedures to better reflect the complexity and evolving nature of real-world GUI environments. We also present a simple and general agent framework, \agent{}, grounded in the principle of critical thinking, which enables agents to identify uncommon states and adapt their plans or actions accordingly.While this work establishes an initial benchmark and baseline, there remain several promising directions for future research. First, we primarily evaluate agents with cheap API models such as Gemini and GPT. Second, \agent{} is designed as a non-training baseline, extending it with learning-based optimization or end-to-end training represents a natural and valuable direction for future work.

\section*{Conclusion}
We introduced \bench{}, a benchmark that evaluates GUI agents under diverse, dynamically perturbed initial states, an aspect largely absent from existing evaluations.  
By constructing task-progress variations via pre-actions, \bench{} enables systematic assessment of robustness and recovery in realistic GUI scenarios. We also presented \agent{}, a lightweight, universal framework that inserts three verification stages into the standard planning loop to improve stability in dynamic environments.  
Our experiments show that current agents, including strong commercial models, still struggle under non-default states, while \agent{} offers consistent gains.

% \newpage

\section*{Impact Statement}

This paper introduces a benchmark and agent framework intended to advance research on dynamic GUI automation. Our goal is to improve the evaluation and reliability of GUI agents in controlled environments, which may support future progress in productivity tools, accessibility assistance, and educational software. As with many advances in machine learning and automation, there are potential risks, including misuse of automated GUI agents for large-scale or unauthorized interactions with software systems. 
Our benchmark does not involve personal or sensitive data and is designed strictly for research use. We encourage practitioners to incorporate appropriate safeguards, such as permission checks, sandboxing, and oversight, when deploying GUI agents in real-world applications. Overall, we believe the broader societal impacts of this work align with those commonly associated with progress in machine learning, and we do not identify specific concerns beyond these general considerations.

%% file: sections/7_append.tex
\section{Task Formulation}
\label{sec:appendix_task_details}

\textbf{GUI Automation.}
Following prior work~\cite{OSWorld, webarena}, we model GUI automation as a partially observable Markov decision process (POMDP) $(\mathcal{S},\mathcal{O},\mathcal{A},\mathcal{T},\mathcal{R})$.  
Given a natural-language query $q$ and optional instructional video $v$, the agent receives an observation $o_t$ of the current state $s_t$ and outputs an action $a_t \in \mathcal{A}$, which updates the environment to $s_{t+1}$.  
The episode terminates when the task succeeds or fails, and $\mathcal{R}$ returns a binary completion signal.

\textbf{\bench{} Tasks.}
As illustrated in Figure~\ref{fig:benchoverview}, each task in \bench{} is instantiated under multiple initial states that all converge to the same goal.  
We create these variants through \emph{pre-actions}—short executable sequences that shift task progress without altering the UI skin, resolution, or layout.  
This design enables controlled task-progress perturbations (Add/Trim/Adjust-step) and complements existing online GUI benchmarks; key differences are summarized in Table~\ref{tab:benchmarkcompare}.

\textbf{Observation Space.}
The agent observes (i) application metadata $m_t$ (e.g., panel structure and parsed UI elements), and (ii) a screenshot $V_t$ of the current GUI state, following ~\cite{assistgui}.  
Metadata provides structured grounding anchors, while screenshots offer holistic visual context for planning and action generation.

\textbf{Action Space.}
Actions include raw mouse and keyboard operations (click, drag, type, shortcuts), with mouse actions parameterized by pixel coordinates in $V_t$.  
We use PyAutoGUI\footnote{\url{https://pyautogui.readthedocs.io}} for execution, and represent each action in the form \texttt{action\_type(arguments)} (Table~\ref{tab:actionspace}).

\begin{table}[h]
    \caption{The action types and examples.}
    \label{tab:actionspace}
    \centering
    \vskip -0.1in
    \resizebox{0.92\linewidth}{!}{
    \begin{tabular}{l|l}
        \toprule
        Action Type & Example \\
        \midrule
        Mouse Movement  & \texttt{moveTo(120, 200)} \\
        Mouse Clicks & \texttt{click(200, 300)} \\
        Keyboard Type & \texttt{write('classes')} \\
        Hotkey & \texttt{hotkey('ctrl', 'a')} \\
        Scrolling & \texttt{scroll(-100)} \\
        Drag & \texttt{dragTo(120, 220, 2)} \\
        Mouse Down and Up & \texttt{mouseDown(); mouseUp()} \\
        Press Keys & \texttt{press('delete')} \\
        Key Down and Up & \texttt{keyDown('shift')} \\
        \bottomrule
    \end{tabular}
    }
    \vskip -0.2in
\end{table}

\input{sections/4_agent}

\begin{table*}[t]
    \centering
    \caption{Performance comparison between \textbf{WorldGUI-Agent} (with Claude-3.5-Sonnet and Claude-Sonnet-4) and \textbf{Agent-S2} (Claude-Sonnet-4). Results are reported across five representative applications.}
    \scalebox{0.80}{
    \begin{tabular}{l|cc|cc|cc|cc|cc}
    \toprule
    \multirow{2}{*}{\textbf{Method}} 
        & \multicolumn{2}{c}{\textbf{PPT}} 
        & \multicolumn{2}{c}{\textbf{VSCode}} 
        & \multicolumn{2}{c}{\textbf{Acrobat}} 
        & \multicolumn{2}{c}{\textbf{VLC}} 
        & \multicolumn{2}{c}{\textbf{File Explorer}} \\
     & Meta & Aug. & Meta & Aug. & Meta & Aug. & Meta & Aug. & Meta & Aug. \\
    \midrule
    Agent-S2 w/ Claude-Sonnet-4 & 45.5 & 18.9 & 45.5 & 37.8 & 18.2 & 14.5 & 20.0 & 16.3 & 60.0 & 64.7 \\
    WorldGUI-Agent w/ Claude-3.5-Sonnet & 54.5 & 39.6 & 54.5 & 26.7 & 63.6 & 20.0 & 50.0 & 18.4 & 50.0 & 17.6 \\
    WorldGUI-Agent w/ Claude-Sonnet-4 & 63.6 & 52.8 & 54.5 & 28.9 & 54.5 & 30.9 & 40.0 & 28.6 & 70.0 & 26.5 \\
    \bottomrule
    \end{tabular}
    }
    \label{tab:comparison_agents_and_worldguiagent}
\end{table*}

\begin{table*}[t]
    \centering
    \caption{Performance comparison between \textbf{WorldGUI-Agent} (with UI-TARS-1.5) and \textbf{UI-TARS-1.5}.}
    \begin{tabular}{l|cc|cc}
    \toprule
    \multirow{2}{*}{\textbf{Method}} 
        & \multicolumn{2}{c}{\textbf{PPT}} 
        & \multicolumn{2}{c}{\textbf{Acrobat}} \\
     & Meta & Aug. & Meta & Aug. \\
    \midrule
    WorldGUI-Agent w/ UI-TARS-1.5 & 36.6 & 18.9 & 36.4 & 9.1 \\
    UI-TARS-1.5 & 27.3 & 17.0 & 9.1 & 1.8 \\
    \bottomrule
    \end{tabular}
    \label{tab:ablateuitars}
\end{table*}

\section{Additional Experiments}
\label{sec:Addtionalexp}

As shown in Table \ref{tab:comparison_agents_and_worldguiagent}, Agent-S2 \citep{agents2} shows competitive results as compared with our \agent{}. We also test on two representative office software to compare the effectiveness of our proposed agentic framework by replacing the base model with UI-TARS-1.5 \citep{uitars} in Table \ref{tab:ablateuitars}. It is noted that to improve the performance of UI-TARS-1.5, we use the GPT-4o to task planning, as we found that UI-TARS struggles with understanding complex desktop software layout and cannot capture the dynamic initial condition changes. We use GPT-4o for better implementation.

\section{Data}
\label{sec:dataappendix}

\subsection{Annotators} In this work, we have four annotators: A, B, C, and D. The team comprises one PhD student, one Master's student, and two undergraduate students. Prior to annotation, all annotators receive training on using the applications in \bench{} to ensure high-quality annotations. For the 10 desktop applications, we divide the software into four parts, assigning each part to a different annotator. For the human tests presented in Table \ref{tab:mainresult}, the annotators demonstrate tasks on software that they did not annotate. As shown in Table \ref{tab:benchmarkcompare}, each annotator is responsible for different software during both the annotation and human testing phases to make the soundness of the Human Test results.

\begin{table*}[h]
    \caption{The annotation arrangement during the annotation and human testing phases by different annotators.}
    \label{tab:annotators}
    \vskip 0.1in
    \centering
    \scalebox{0.9}{
    \begin{tabular}{c|cc}
    \toprule
       Annotators  & Annotation Phase & Human Test Phase \\
    \midrule
        A & PowerPoint, Word, Excel & VSCode, VLC Player, Web\\
        B & Adobe Acrobat, VLC Player & Excel, Settings\\
        C & Settings, Web & PowerPoint, File Explorer, Youtube\\
        D & VSCode, File Explorer & Word, Adobe Acrobat\\
    \bottomrule
    \end{tabular}
    }
    
\end{table*}

\begin{figure*}[t]
\centering
\includegraphics[width=0.9\linewidth]{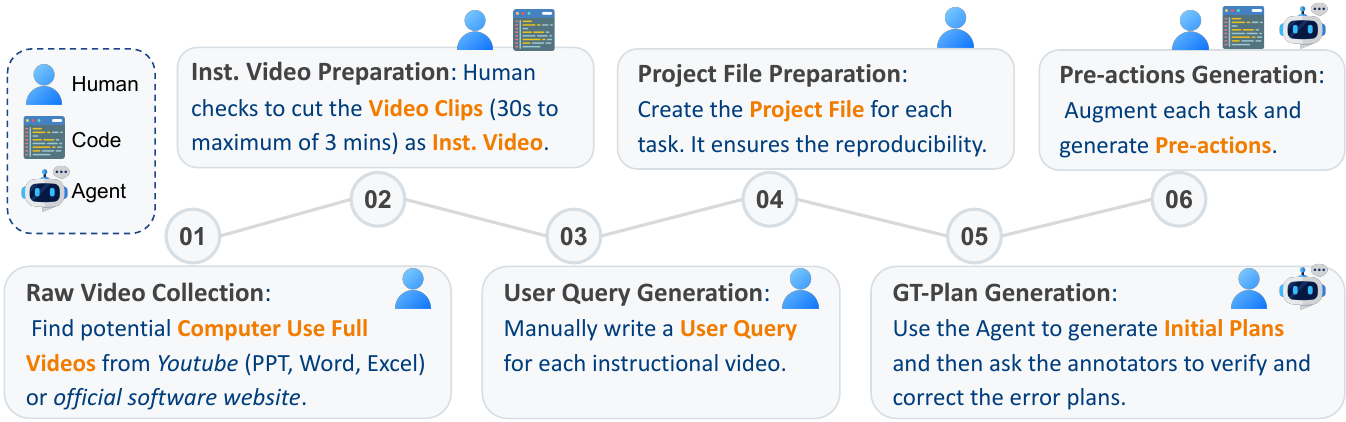}
\caption{Pipeline of Data Construction. Human: Represents the annotators. Code: Refers to the scripts (e.g., Python Code) utilized to achieve the goal. Agent: We design an agent built upon the MLLMs to achieve the goal.}
\label{fig:datapipeline}
\end{figure*}

\subsection{Creating Augmented Tasks} In our study, to simulate dynamic testing processing in real GUI interactions, we propose to design GUI tasks with various initial tasks. Specifically, we propose pre-actions before executing the task. The pre-actions primarily serve two purposes: \textbf{1) Simulating Intermediate Task States}: Pre-actions can complete specific steps of a task, creating a starting point from an intermediate state. This approach addresses scenarios where users may seek AI assistance because they are unable to complete a task. For example, if the task involves opening a dropdown menu, the pre-action may pre-open the menu. If the agent fails to recognize this precondition and follows its plan to click the menu again, it might inadvertently close the menu, causing task failure.

\subsection{Introducing Diverse Initial Context States} 
\begin{figure*}[h]
\centering
\includegraphics[width=\linewidth]{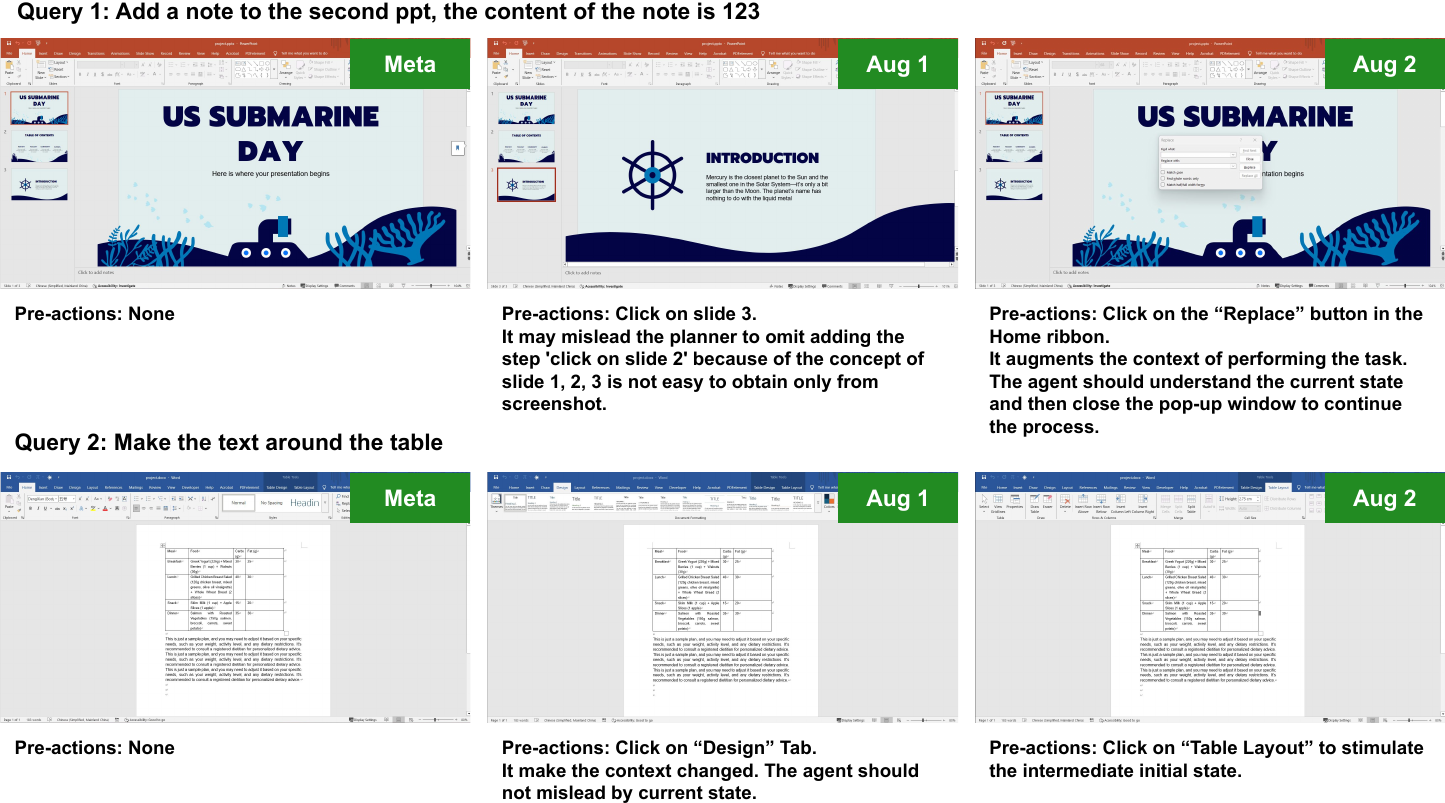}
\caption{We present the examples of conducting the augmentations from the meta task.}
\label{fig:augexample}
% \vskip -0.2in
\end{figure*}

Pre-actions can also introduce variations in the initial state, such as opening random tabs or settings. This ensures that the starting state is unconventional, challenging the agent to adapt by modifying its plan or adding new steps. We illustrate one example in Figure~\ref{fig:dataaugment}. Here, the meta task and augmented task, have the same user query and instructional video and it will ideally have the same final state. We additionally provide more examples about augmenting the meta task in Figure \ref{fig:augexample}.

\section{Detailed Experimental Results}

Table \ref{tab:windowagentarena_details} shows the detailed results of \agent{} across individual applications in WindowsAgentArena \citep{windowsagentarena} benchmark. The results of this related Windows-centric interactive GUI benchmark indicate that current the desktop GUI tasks are more challenging than web tasks. As we complete 11 out of 17 tasks in Web Browsing, a similar phenomenon is also discovered in Table 4. 

\begin{table*}[t]
    \centering
    \caption{Detailed experimental results of \agent{} across individual applications in WindowsAgentArena \citep{windowsagentarena}.}
    \begin{tabular}{llccc}
    \toprule
    \textbf{Domain} & \textbf{Application} & \textbf{\#Tasks} & \textbf{\#Successes} & \textbf{SR (\%)} \\
    \midrule
    Web Browsing & chrome & 17 & 11 & 64.71 \\
    Windows Utilities & clock & 4 & 2 & 50.00 \\
    Windows System & file\_explorer & 19 & 7 & 36.84 \\
    Office & libreoffice\_calc & 24 & 1 & 4.17 \\
    Office & libreoffice\_writer & 19 & 2 & 10.53 \\
    Windows Utilities & microsoft\_paint & 3 & 1 & 33.33 \\
    Web Browsing & msedge & 13 & 5 & 38.46 \\
    Windows Utilities & notepad & 2 & 1 & 50.00 \\
    Windows System & settings & 5 & 4 & 80.00 \\
    Media \& Video & vlc & 21 & 6 & 28.57 \\
    Coding & vs\_code & 24 & 8 & 33.33 \\
    Windows Utilities & windows\_calc & 3 & 0 & 0.00 \\
    \midrule
    \textbf{Overall} & & \textbf{154} & \textbf{48} & \textbf{31.17} \\
    \bottomrule
    \end{tabular}
    \label{tab:windowagentarena_details}
\end{table*}

\section{Computational Costs Discussion}
The average number of execution steps and tokens consumed are shown below Table \ref{tab:tokencosts}. The execution steps are calculated based on our experimental log files, while the token costs are sampled from representative tasks in each category by taking Actor module as an example.

\begin{table*}[htbp]
  \centering
  \caption{Average execution steps and token costs on different software.}
  \scalebox{0.76}{
  \begin{tabular}{lccc}
    \toprule
    Application category & Average execution steps & Input tokens per step (Actor) per task & Output tokens per step (Actor) per task \\
    \midrule
    Office         & $\sim$23 & 2350 & 212 \\
    Win.\ Usage    & $\sim$20 & 1929 & 108 \\
    Web            & $\sim$17 & 1637 & 84  \\
    \bottomrule
  \end{tabular}}
\label{tab:tokencosts}
\end{table*}

Take a Windows Setting task as an example, we provide detailed time costs across different modules tested on a desktop PC with AMD Ryzen 7 5800H CPU. Task length: 6 (generated by Planner+Planner-Critic). To facilitate a fair comparison, we additionally selected two of the latest SOTA agents, Agent-S (2024-10-08) and Agent-S2 (2025-04-01), and measured their run times on the same successful task under identical hardware and the same base MLLM (Claude-Sonnet-4). The results are shown in Tables~\ref{tab:agent-runtime}, \ref{tab:agent-s-runtime}, \ref{tab:agent-s2-runtime}. To summarize, our \textbf{WorldGUI-Agent} shows a competitive running time of 129.55s, as compared with Agent-S (131.98s) and Agent-S2 (108.64s). The main computational costs of our designed modules are largely affected by the underlying large multimodal model, leaving room for acceleration optimization.

Since desktop GUI automation is still in its early stages, such computational costs are currently unavoidable. For reference, even OpenAI’s Deep Research reportedly takes over 10 minutes in daily usage. According to OpenAI Operator’s report, achieving 38.1\% on OS-World requires over 100 steps, which is similarly costly. In summary, there remains a clear tradeoff between performance and time costs in GUI automation, and this challenge is shared across the community.

\begin{table*}[htbp]
  \centering
  \caption{Running time with \textbf{WorldGUI-Agent (ours)}.}
  \begin{tabular}{clc}
    \toprule
    Subtask Index & Executed Modules & Time (seconds) \\
    \midrule
    0 & Planner        & 4.48 \\
    0 & Planner-Critic & 11.47 \\
    1 & Parser         & 2.03 \\
    1 & Step-Check     & 4.95 \\
    1 & Actor          & 7.47 \\
    1 & Parser         & 2.07 \\
    1 & Actor-Critic   & 7.38 \\
    2 & Parser         & 2.03 \\
    2 & Step-Check     & 6.71 \\
    2 & Actor          & 6.05 \\
    2 & Parser         & 2.07 \\
    2 & Actor-Critic   & 7.97 \\
    3 & Parser         & 2.12 \\
    3 & Step-Check     & 6.35 \\
    3 & Actor          & 6.71 \\
    3 & Parser         & 2.22 \\
    3 & Actor-Critic   & 10.06 \\
    4 & Parser         & 2.01 \\
    4 & Step-Check     & 6.19 \\
    4 & Actor          & 8.54 \\
    4 & Parser         & 2.40 \\
    4 & Actor-Critic   & 9.55 \\
    5 & Parser         & 2.22 \\
    5 & Step-Check     & 6.50 \\
    \midrule
    Total              & - & 129.55 \\
    Average (per action) & - & 22.72 \\
    \bottomrule
  \end{tabular}
  \label{tab:agent-runtime}
\end{table*}

\begin{table*}[htbp]
  \centering
  \caption{Running time with \textbf{Agent-S}.}
  \begin{tabular}{clc}
    \toprule
    Step & Executed Modules & Time (seconds) \\
    \midrule
    0 & Manager & 2.02 \\
    1 & Manager & 10.11 \\
    2 & Worker  & 16.05 \\
    3 & Worker  & 12.38 \\
    4 & Worker  & 17.30 \\
    5 & Worker  & 12.26 \\
    6 & Worker  & 15.17 \\
    7 & Worker  & 15.46 \\
    8 & Worker  & 11.83 \\
    9 & Worker  & 19.40 \\
    \midrule
    Total              & - & 131.98 \\
    Average (per worker) & - & 14.98 \\
    \bottomrule
  \end{tabular}
  \label{tab:agent-s-runtime}
\end{table*}

\begin{table*}[htbp]
  \centering
  \caption{Running time with \textbf{Agent-S2}.}
  \begin{tabular}{clc}
    \toprule
    Step & Executed Modules & Time (seconds) \\
    \midrule
    0 & Manager & 11.87 \\
    1 & Worker  & 7.27 \\
    2 & Worker  & 13.21 \\
    3 & Worker  & 13.59 \\
    4 & Worker  & 14.04 \\
    5 & Manager & 10.61 \\
    6 & Worker  & 9.87 \\
    7 & Manager & 7.53 \\
    8 & Worker  & 20.65 \\
    \midrule
    Total              & - & 108.64 \\
    Average (per worker) & - & 13.22 \\
    \bottomrule
  \end{tabular}
  \label{tab:agent-s2-runtime}
\end{table*}

\section{Examples of Augmentations}

\begin{figure*}[ht]
    \centering
    \includegraphics[width=\linewidth]{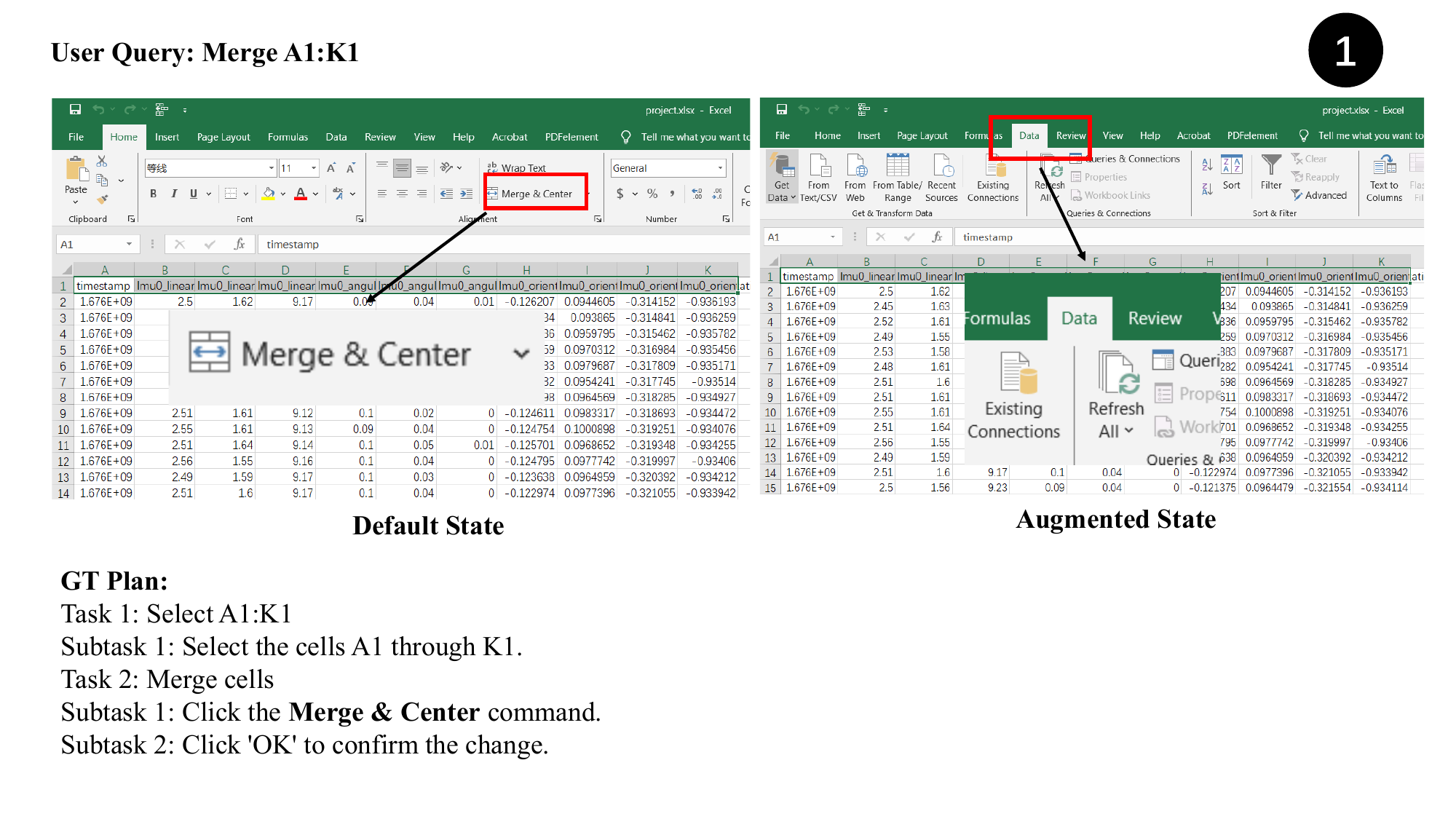}
    \caption{Augmented example of an Excel Task.}
    \label{fig:excelaug1}
\end{figure*}

\begin{figure*}[ht]
    \centering
    \includegraphics[width=\linewidth]{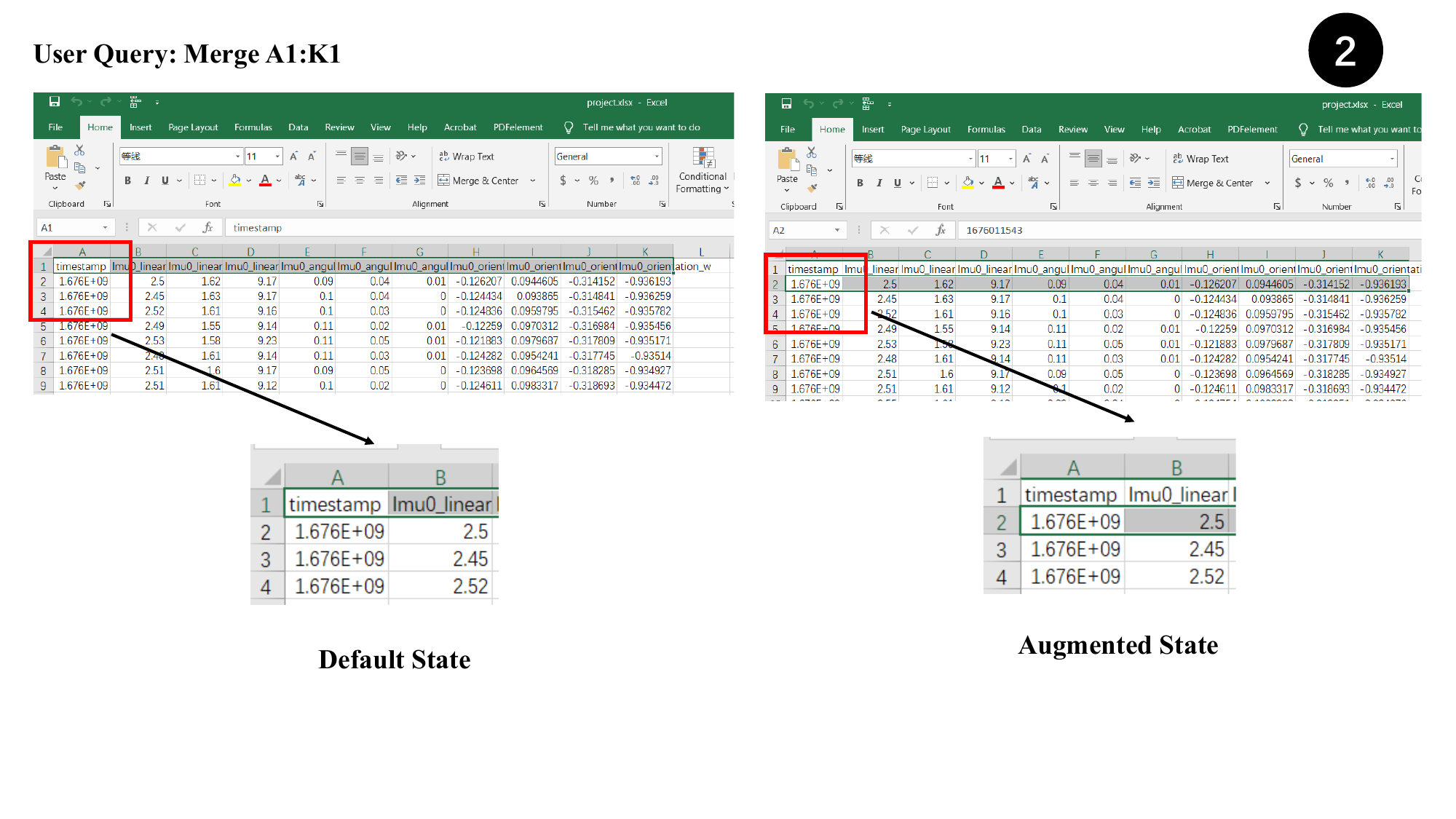}
    \caption{Augmented example of an Excel Task.}
    \label{fig:excelaug2}
\end{figure*}

\begin{figure*}[ht]
    \centering
    \includegraphics[width=\linewidth]{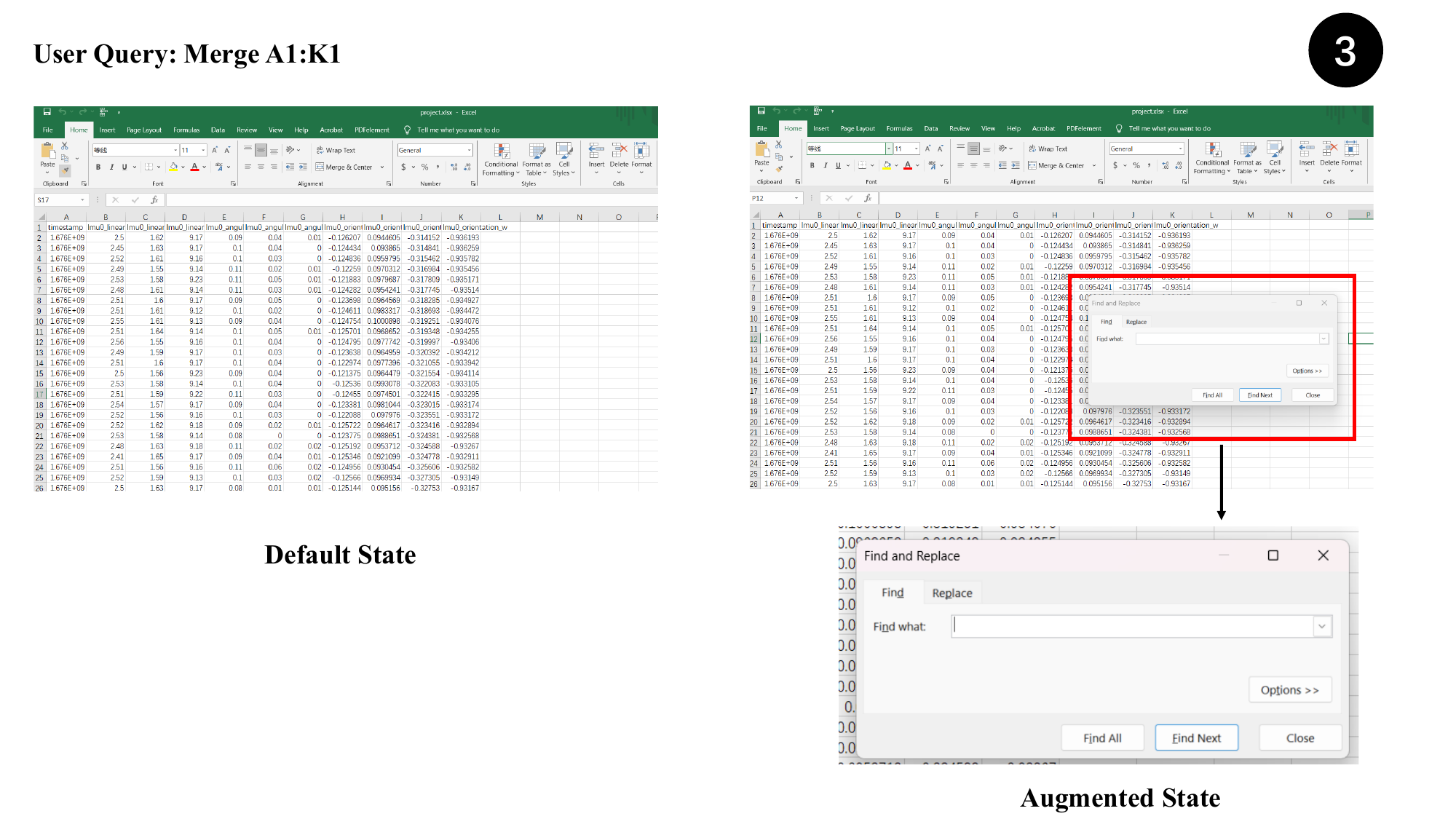}
    \caption{Augmented example of an Excel Task.}
    \label{fig:excelaug3}
\end{figure*}

\begin{figure*}[ht]
    \centering
    \includegraphics[width=\linewidth]{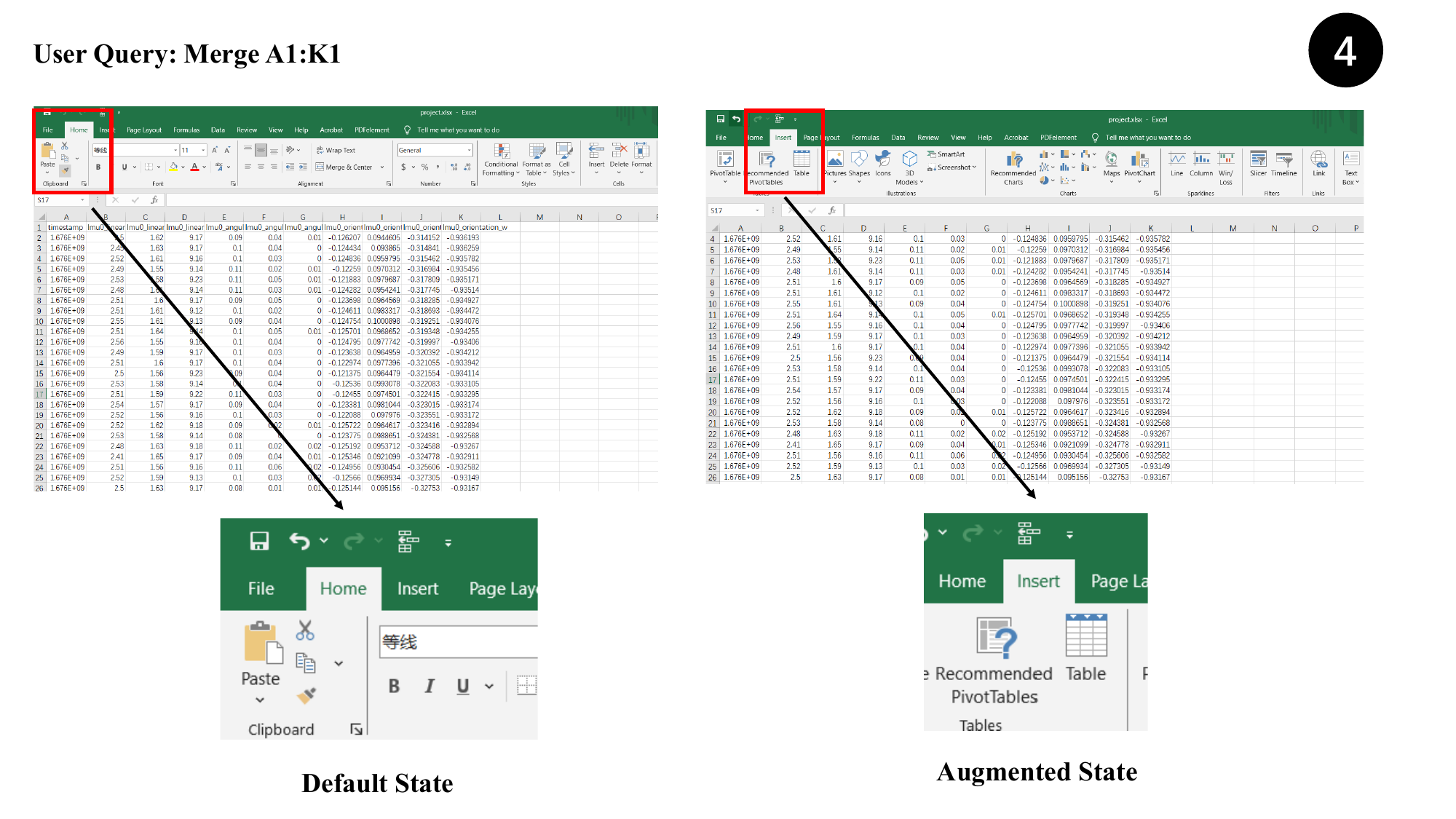}
    \caption{Augmented example of an Excel Task.}
    \label{fig:excelaug4}
\end{figure*}

\begin{figure*}[ht]
    \centering
    \includegraphics[width=\linewidth]{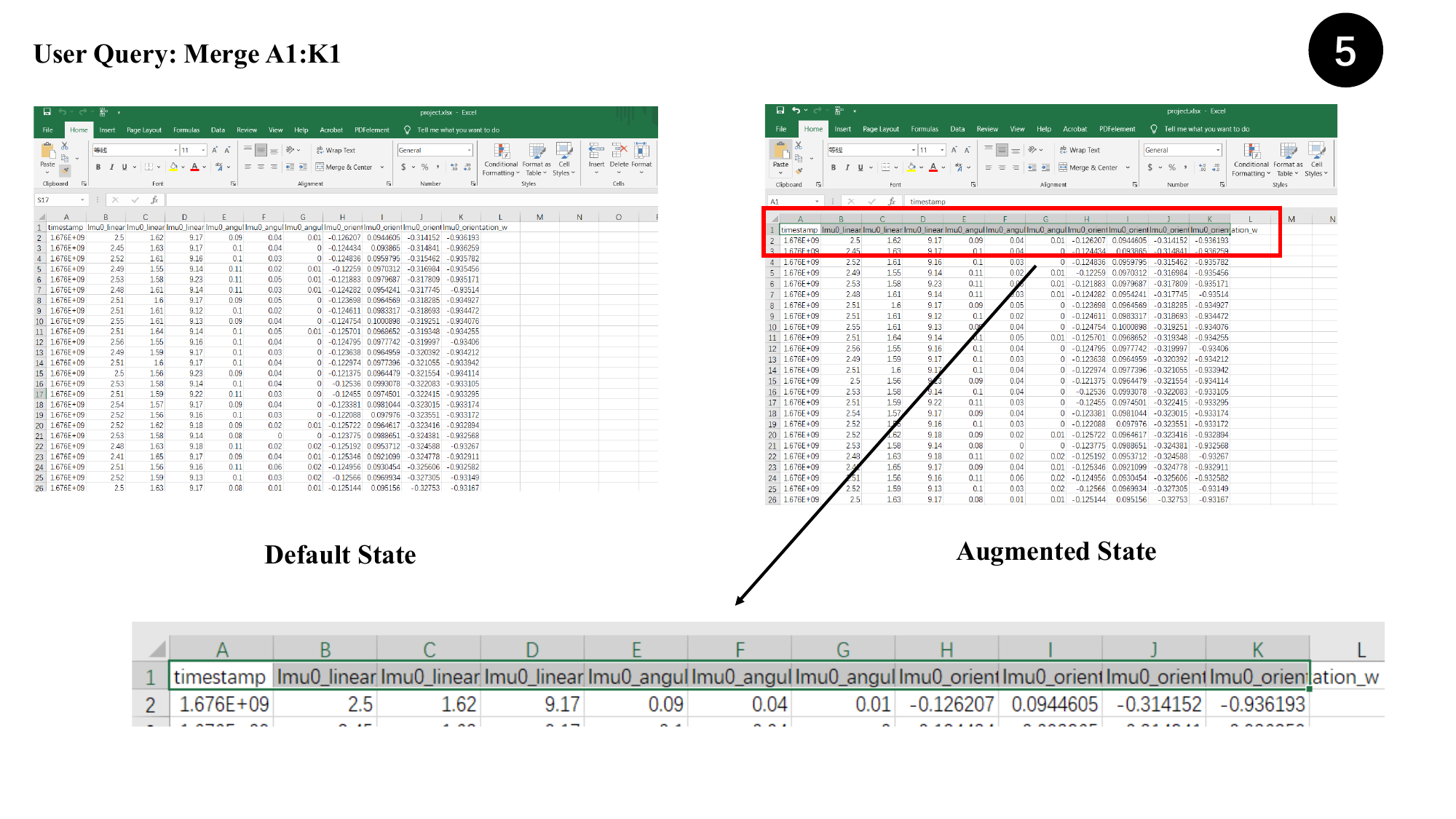}
    \caption{Augmented example of an Excel Task.}
    \label{fig:excelaug5}
\end{figure*}

\begin{figure*}[ht]
    \centering
    \includegraphics[width=\linewidth]{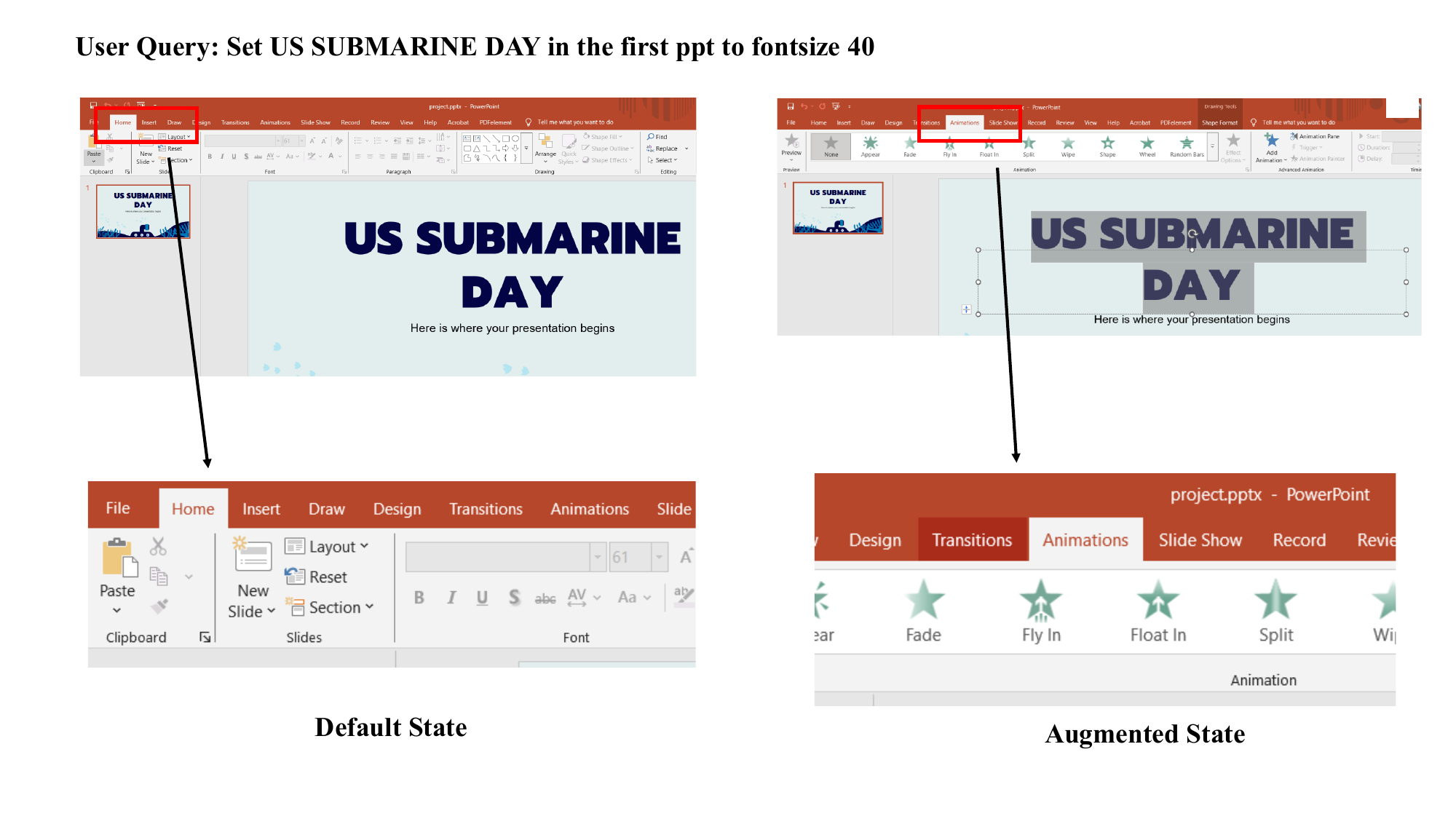}
    \caption{Augmented example of a PowerPoint Task.}
    \label{fig:pptaug1}
\end{figure*}

\begin{figure*}[ht]
    \centering
    \includegraphics[width=\linewidth]{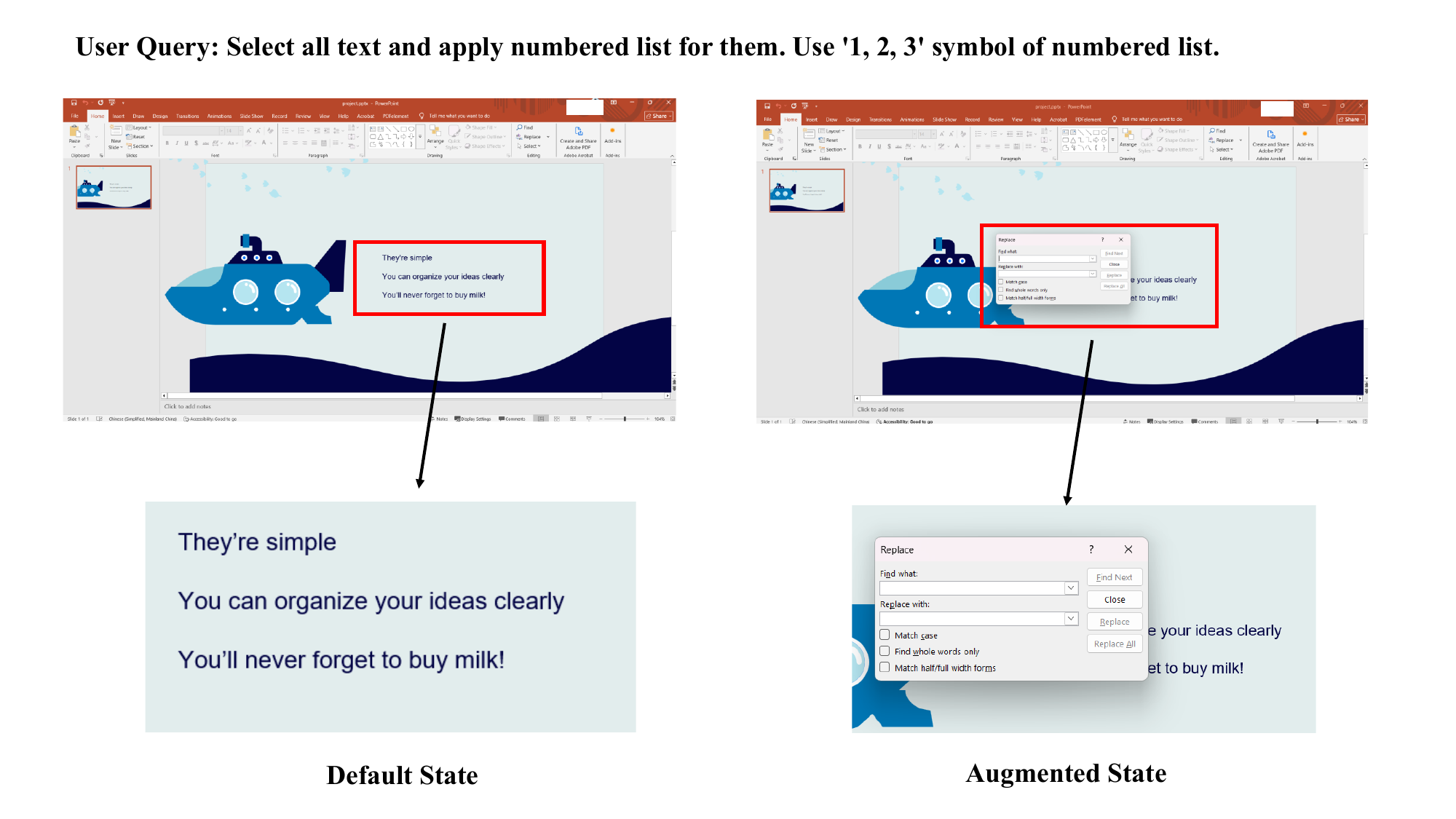}
    \caption{Augmented example of a PowerPoint Task.}
    \label{fig:pptaug2}
\end{figure*}

In this section, we present several augmentation examples in Figures \ref{fig:excelaug1}, \ref{fig:excelaug2}, \ref{fig:excelaug3}, \ref{fig:excelaug4}, \ref{fig:excelaug5}, \ref{fig:pptaug1}, \ref{fig:pptaug2}. It is noted that our augmentations are not only making the first step changing but also require the agent add new step in its second step. For instance, in Figure \ref{fig:excelaug1}, our augmentation is about click on Data tab in the ribbon, in the default software state, the Merge \& Center button exhibit in the Home tab, there is no need to click on Home tab, after our augmentations, the agent should add a new task ``Click on Home Tab'' before it click on the Merge \& Center button. Similarly, in Figure \ref{fig:pptaug1}, the text editing buttons are under the Home Tab, if we augment the initial state with other Tab like Animation Tab, after the first step ``Select the text 'US SUBMARINE DAY' '', the agent should add a new step like ``Click on Home Tab'' back to the default state for task execution. Except for adding new steps, we also present an example about adjust the step in Figure \ref{fig:excelaug2}, as the target is about merging cells A1 to K1, we augment the initial state by selecting A2 to K2. Such a slight difference may mislead the agent to perceive such a minor difference, and the agent may jump the first step about selecting the correct cells lead to finally unsucess. In Figure \ref{fig:excelaug3} and Figure \ref{fig:pptaug2}, we show two examples of introducing pop-up window in the initial state which require the agents accurately identify the pop-up windows and correctly close it by replanning the task based on the visual screenshot not only strictly planning based on inherited knowledge or the instructinal videos. In Figure \ref{fig:excelaug4}, we show an example of changing the interface by clicking the Data tab to hide the Merge \& Center button under the Home tab. In Figure \ref{fig:excelaug5}, we complete the first step about selecting A1 to K1, which requires the agent to jump this step to reduce the time costs.

\section{\agent{} Reasoning Loop Algorithm}
In this section, we provide the details of our reasoning loop algorithm in Algorithm \ref{alg:reasoningloop}.

\begin{algorithm*}[h]
   \caption{\agent{} Reasoning Loop Algorithm}
   \label{alg:reasoningloop}
\begin{algorithmic}
   \STATE {\bfseries Input:} State $s$, Action Code $C$, Screenshot $V$, Metadata $m$, Current subtask $S$, Critic\_count $z$
   \STATE Generate task plan $p$ with \textbf{Planner} and \textbf{Planner-Critic}
   \STATE Initial current subtask $S_{t=0} = S_1^1$, where $S_1^0$ is the $1$-th subtask in the $1$-th milestone of $p$. 
   \STATE Initial $s_0 = \textless Continue\textgreater$  
   \WHILE{$S_t$ is not end and $t < $ max trials}
   \STATE Observe metadata $m_t$ and Screenshot $V_t$ from Env.
    \STATE Obtain state $s_t$ by running \textbf{Step-Check}.
    \IF{$s_t = \textless Next\textgreater$}
    \STATE Go to the next task $S_{t+1}=next(S_{t})$
    \ENDIF
    \STATE Check potential modification of subtask $S_t$ 
    \STATE Obtain action code $C_t$ by running \textbf{Actor};
     Execute the action code $C_t$ in the Env.; Observe metadata $m_t$ and Screenshot $V_t$ from Env.
    \STATE Set $C_t=$ \text{None}; $t = t+1$; Set state $s_t = \textless Critic\textgreater$ (For each subtask, the first step is finished, then execute the actor-critic process)
    \STATE Observe metadata $m_t$ and Screenshot $V_t$ from Env.
    \STATE Running \textbf{Actor-Critic} and obtain the state $s_t$
    \IF{$s_t = \textless Next\textgreater$}
    \STATE Go to the next task $S_{t+1}=next(S_{t})$.
    \ENDIF
    \WHILE{$s_t = \textless Critic\textgreater$ and $z <$ max critique trials}
    \STATE Running \textbf{Actor-Critic} and obtain the state $s_t$ and corrected action code $C_t$
    \IF{$s_t = \textless Next\textgreater$}
    \STATE Go to the next task $S_{t+1}=next(S_{t})$.
    \ENDIF
    \STATE Execute the action code $C_t$ in the Env.; Observe metadata $m_t$ and Screenshot $V_t$ from Env.
    \STATE Set $C_t=$ \text{None}; $z=z+1$
    \ENDWHILE
   \STATE Go to the next task $S_{t+1}=next(S_{t})$
   \STATE $t = t+1$
   
   \ENDWHILE
\end{algorithmic}
\end{algorithm*}

\section{Qualitative Results}
\label{sec:qualitative_results}

\begin{figure*} [h]
\centering
\includegraphics[width=\linewidth]{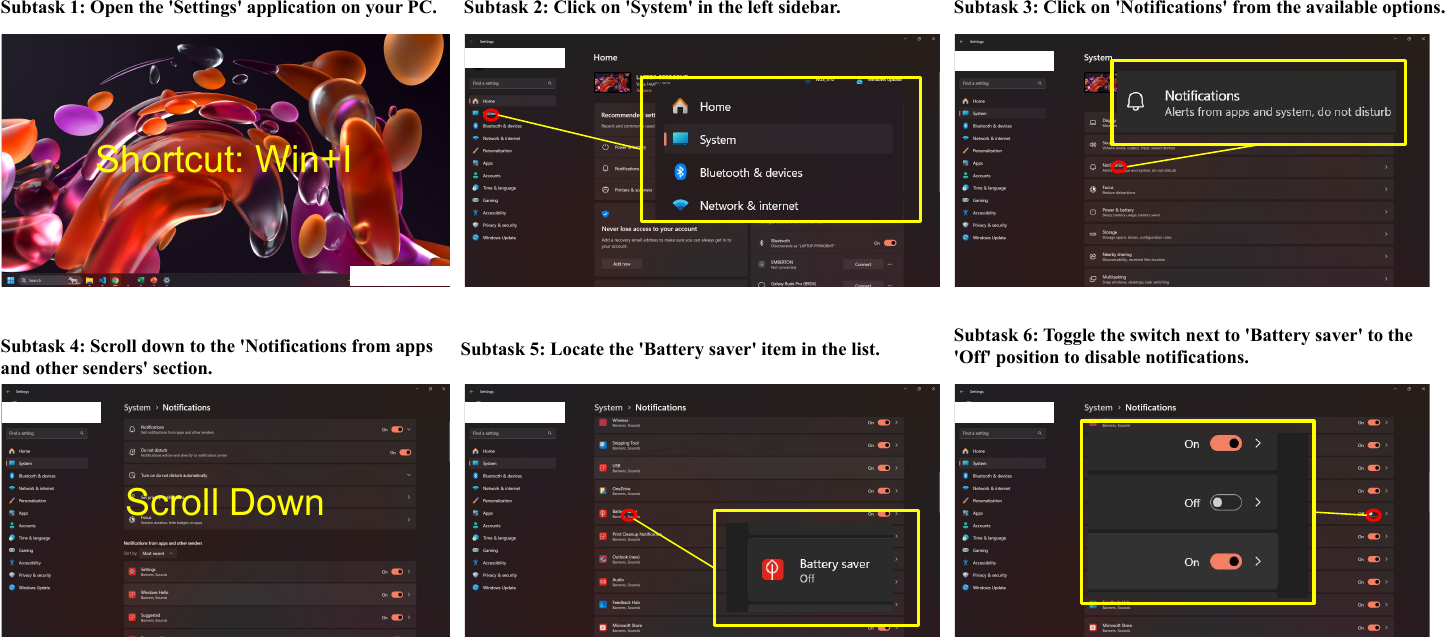}
\vskip -0.1in
\caption{We show one successful prediction of our \agent{}.}
\label{fig:ansuccessexample}
\end{figure*}

\begin{figure*} [h]
\centering
\includegraphics[width=\linewidth]{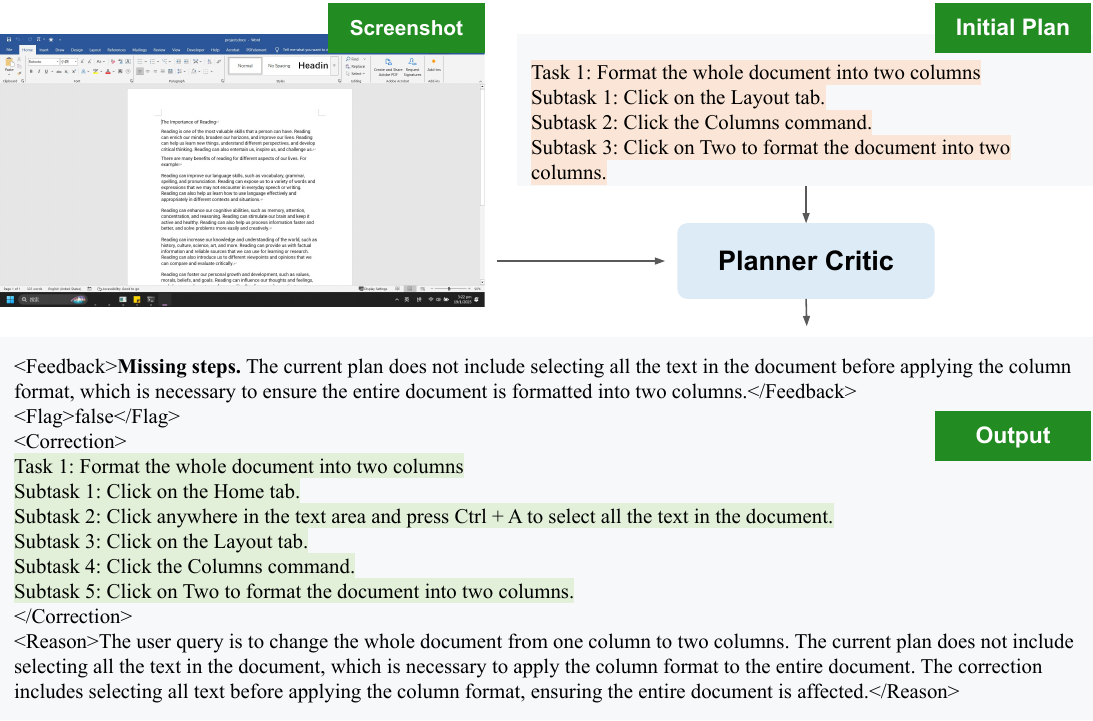}
\vskip -0.1in
\caption{An example of using Planner-Critic to correct the plan.}
\label{fig:plannercriticexample}
\end{figure*}

\begin{figure*} [h]
\centering
\includegraphics[width=\linewidth]{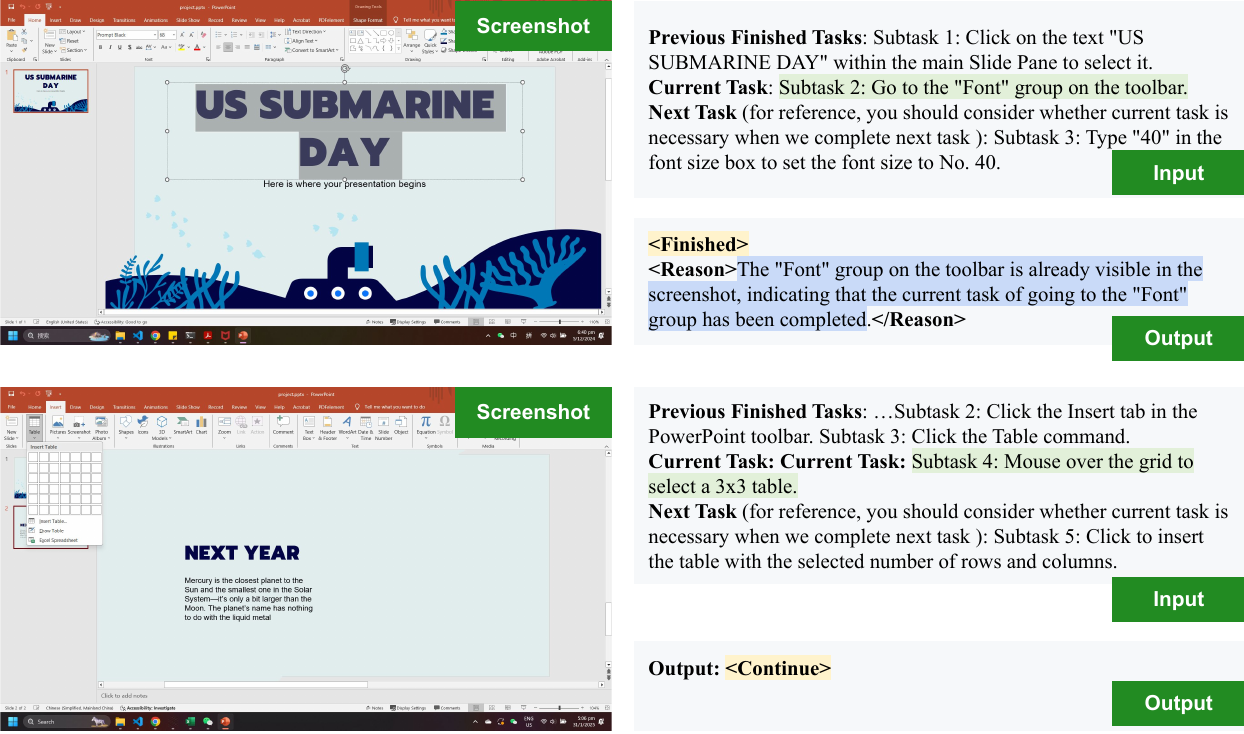}
\vskip -0.1in
\caption{Two examples of using Step-Check to check the subtask status.}
\label{fig:stepcheckexample}
\end{figure*}

\begin{figure*} [h]
\centering
\includegraphics[width=\linewidth]{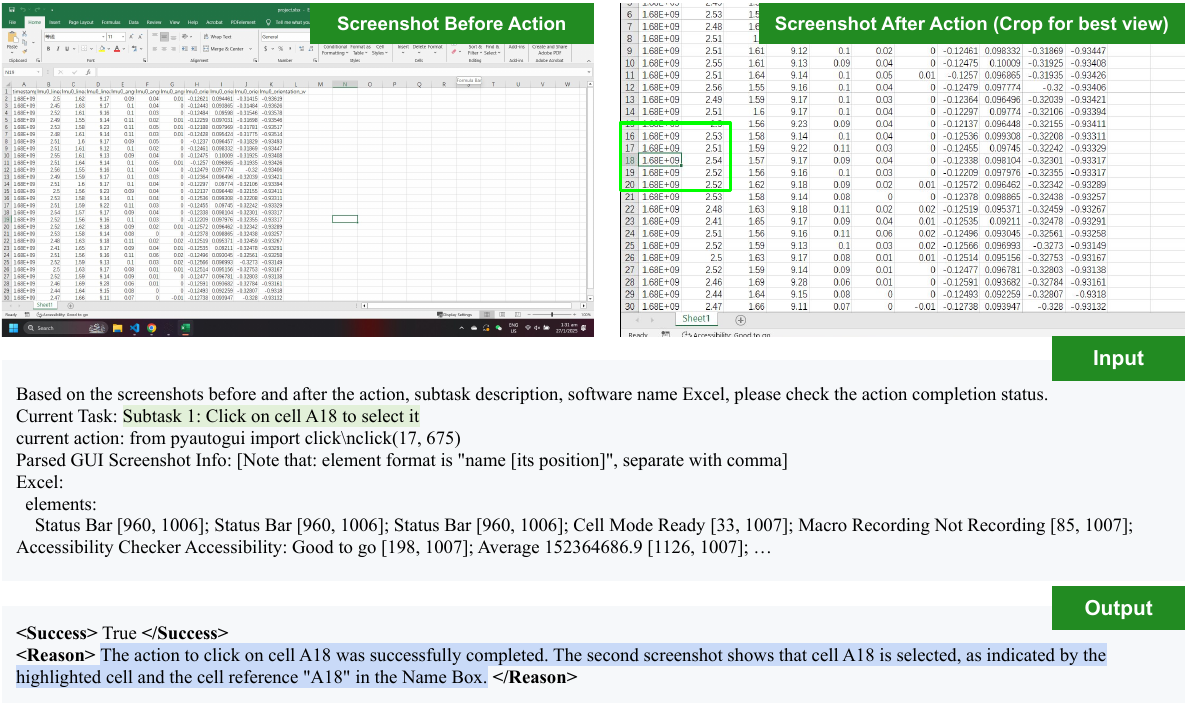}
\vskip -0.1in
\caption{An example of using Actor-Critic to correct the actions.}
\label{fig:actorcriticexample}
\end{figure*}

\begin{figure*} [h]
\centering
\includegraphics[width=\linewidth]{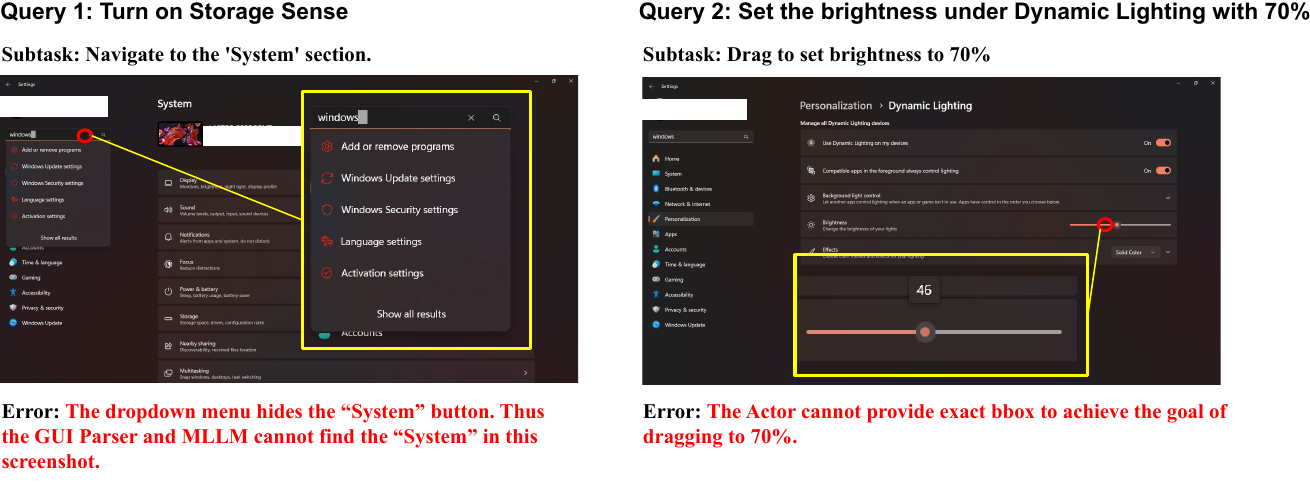}
\caption{We display some common errors.}
\label{fig:failurecases1}
\end{figure*}

\begin{figure*} [htp]
\centering
\includegraphics[width=\linewidth]{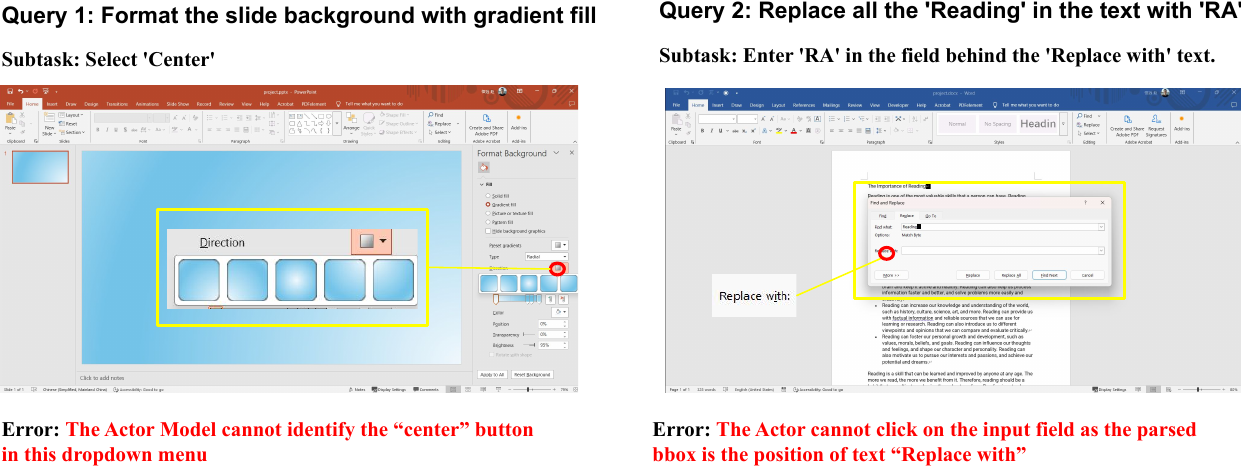}
\caption{We display some common errors}
\label{fig:failurecases2}
\end{figure*}

(1) In Figure \ref{fig:ansuccessexample}, we present a successful prediction example, demonstrating that the \bench{} can effectively plan each step for a task, accurately perceive specific elements in the GUI, and convert them into the correct action code. Additionally, we display the parsed GUI elements, which can accurately identify most content, including small icons and dense text elements.
(2) We provide the visualization results of using Planner-Critic, Step-Check, and Actor-Critic in Figure \ref{fig:plannercriticexample}, Figure \ref{fig:stepcheckexample}, and Figure \ref{fig:actorcriticexample}. These qualitative results demonstrate the effectiveness of these critical modules in GUI automation.
(3) We also highlight some common errors encountered. 1)
The model has difficulty obtaining the desired information when we augment the task by invoking the dropdown menu of the Settings application. As shown on the left of Figure \ref{fig:failurecases1}, when we click on the 'System' button on the left, it is challenging for our model to extract the button's position as it is hidden. Such cases require the model to have a higher level of ability to delete the content in the input box or click on the blank area. 2) As shown in the right of Figure \ref{fig:failurecases1}, the model has difficulty dragging a bar to achieve the desired value. 3) The model struggles with the visual choice when there is no text information in the screenshot, as shown on the left of Figure \ref{fig:failurecases2}. The subtask aims to select the center button, but the current model makes it hard to detect the center choice only from the screenshot. 4) The model cannot successfully locate the position of the input box, as the GUI parser will easily locate the text location 'Replace with', it always outputs the action like clicking on the 'Replace with', which will destroy the whole task's success.

\begin{figure*} [h]
    \centering
    \includegraphics[width=0.48\linewidth]{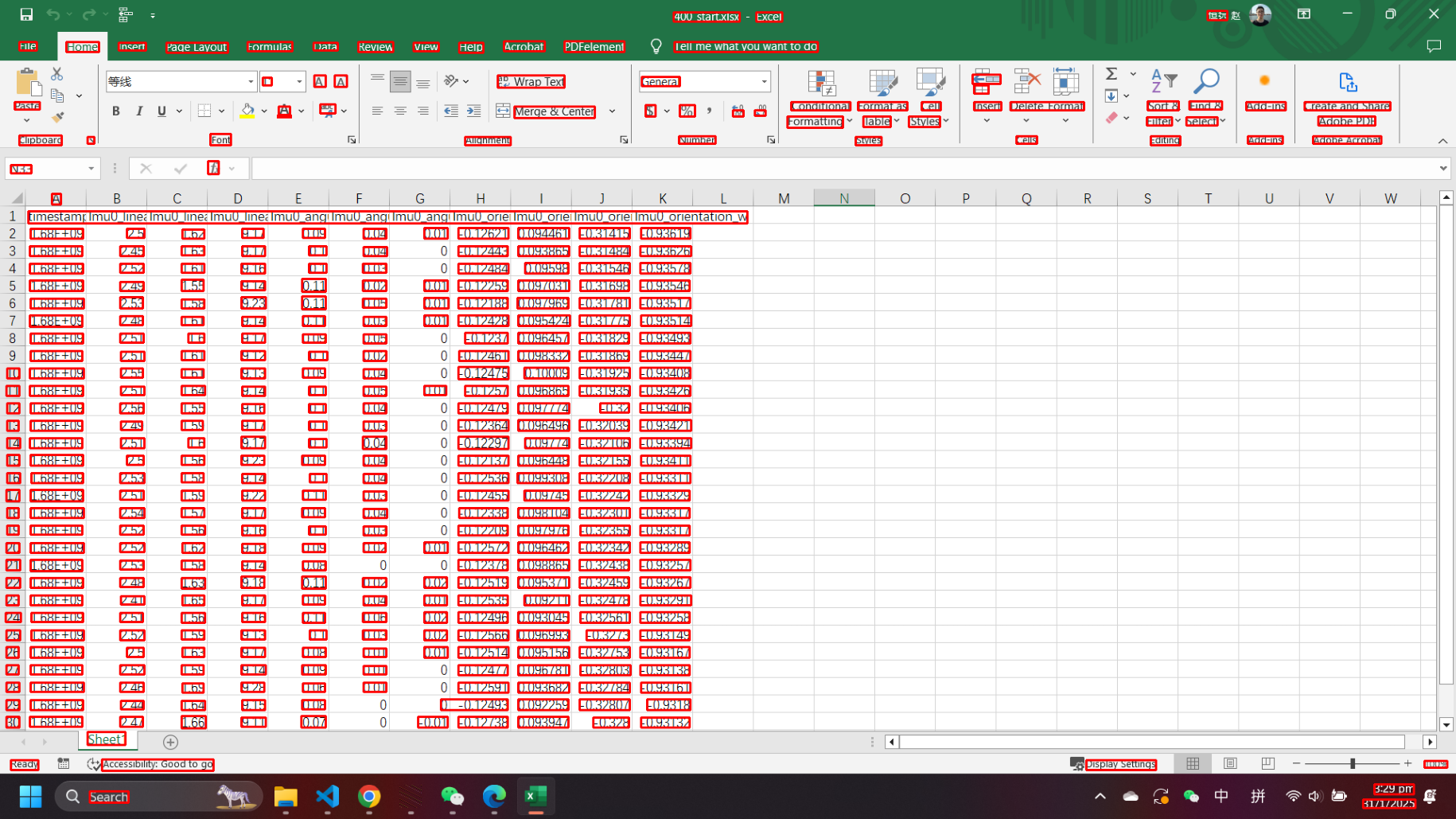}
    \includegraphics[width=0.48\linewidth]{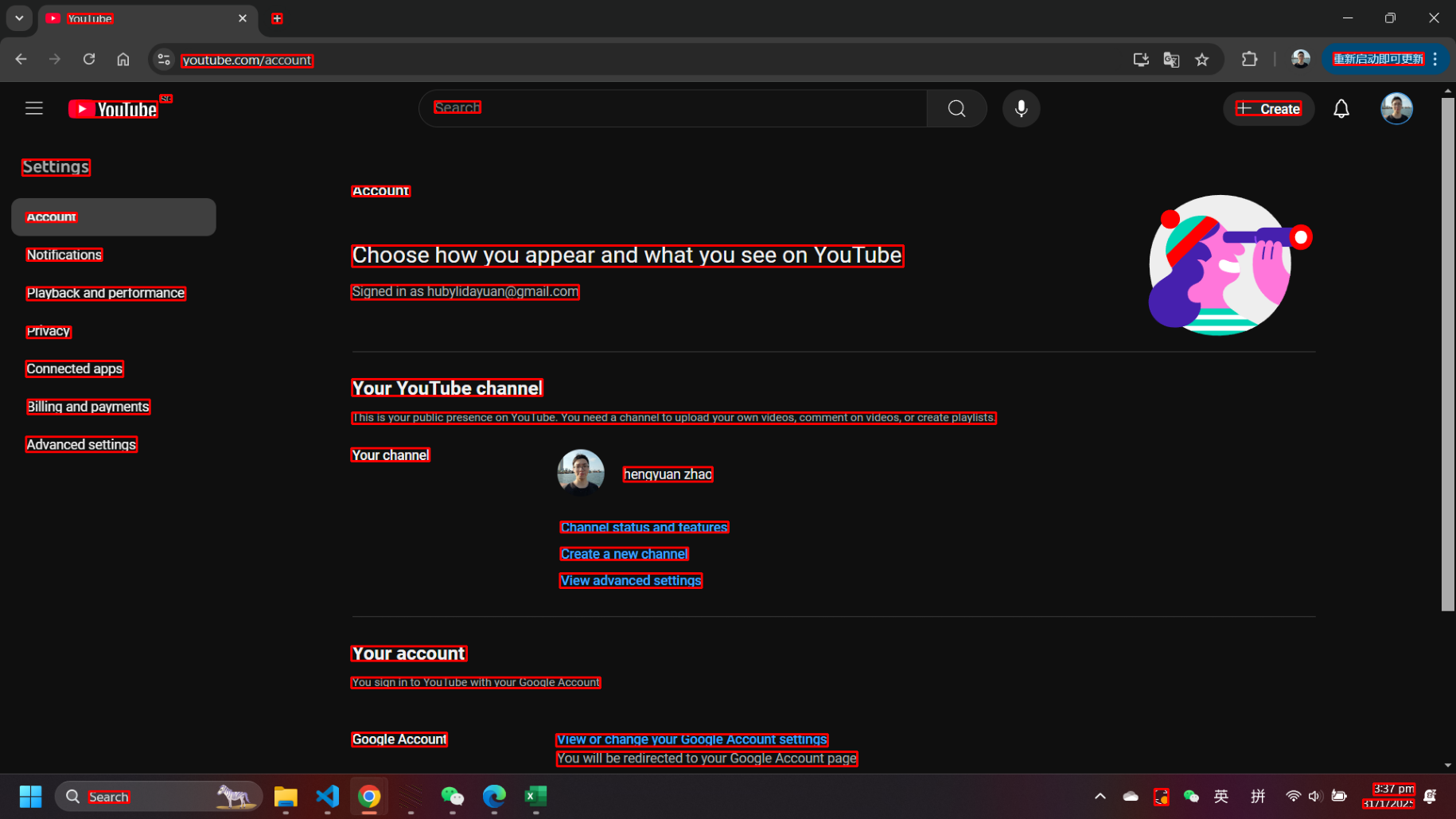}
    \vskip -0.1in
    \caption{We show two examples of using GUI Parser to obtain the element position information.}
    \label{fig:parservisuall}
\end{figure*}

%% file: sections/4_agent.tex
\section{\agent{}: Thinking before Doing}
\label{sec:agent}

\begin{figure*}[t]
\vskip -0.1in
\centering
\includegraphics[width=0.82\linewidth]{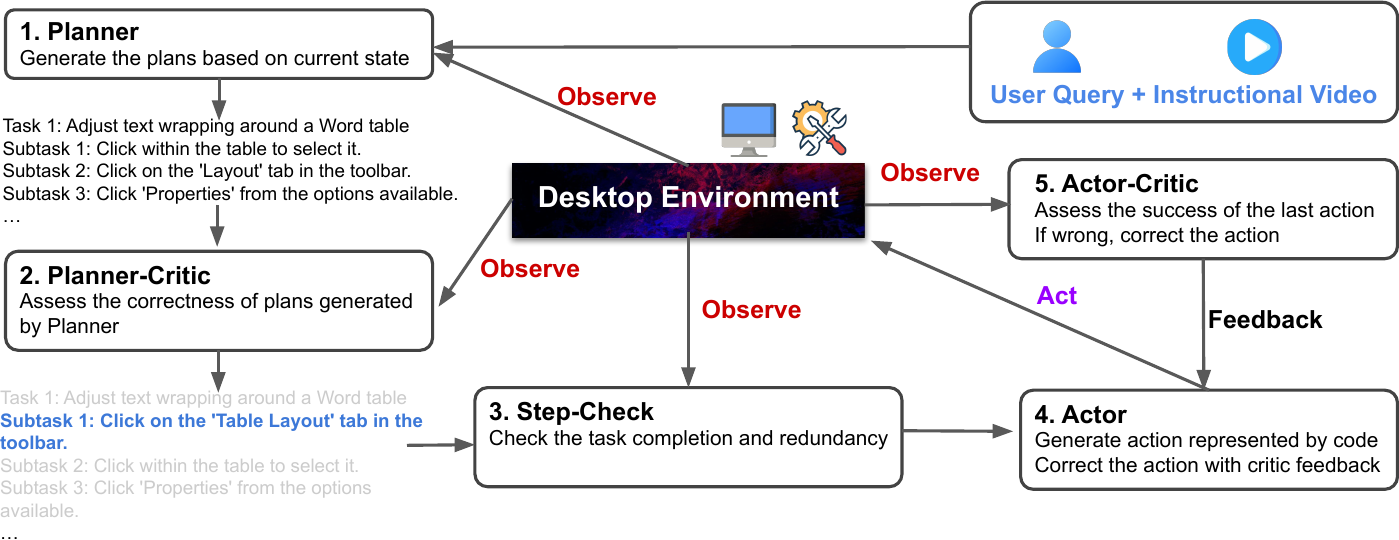}
\vskip -0.1in
\caption{\textbf{\agent{}}. The Planner module receives the user query and an instructional video as input and generates an initial plan. This plan is then refined and executed step by step. Before each step is passed to the Actor module, it undergoes a Step-Check. After the Actor produces an action, the Actor-Critic module iteratively verifies the completion of the action and makes corrections.}
\label{fig:agentoverview}

\end{figure*}

% \begin{table*}[h]
%   \centering
%   \caption{Comparison with other closely related agents. Most existing agents solely focus on post-action evaluation but omit the post-planning critique and pre-action validation in handling dynamic GUI environments.}
%   \label{tab:compareagents}
%   \scalebox{0.8}{
%   \begin{tabular}{lccc}
%     \toprule
%     Method & Post-Planning Critique & Pre-Action Validation & Post-Action Evaluation \\
%     \midrule
%     Mobile-Agent \citep{wang2024mobile}         & \xmark{}  & \xmark{}  & \cmark{} \\
%     Mobile-Agent-V2 \citep{wang2024mobile}      & \xmark{}  & \xmark{}  & \cmark{} \\
%     AssistGUI \citep{assistgui}   & \xmark{} & \xmark{}  & \cmark{} \\
%     Agent-S  \citep{agents1}            & \xmark{} & \xmark{} & \cmark{} \\
%     Mobile-Agent-E  \citep{wang2025mobileagentE}     & \xmark{} & \xmark{} & \cmark{} \\
%     \agent{} (ours)  & \cmark{} & \cmark{} & \cmark{} \\
%     \bottomrule
%   \end{tabular}
%   }
% \end{table*}

In this section, we introduce an universal GUI framework \textbf{\agent{}} with a core and essential designing principle: \textit{critical thinking}, which is vital for designing GUI agents capable of handling dynamic environments that have been overlooked in prior GUI agents ~\citep{hong2024cogagent, cheng2024seeclick, lin2024showui, zhang2023appagent, agents1}. The \agent{} includes the \textbf{five fundamental but essential components} as in Figure \ref{fig:agentoverview} and an \textbf{Interaction reasoning loop} detailed in Algorithm \ref{alg:reasoningloop}. 
We summarize our critical designs in the following:

\begin{figure}[t]

\centering
\includegraphics[width=0.9\linewidth]{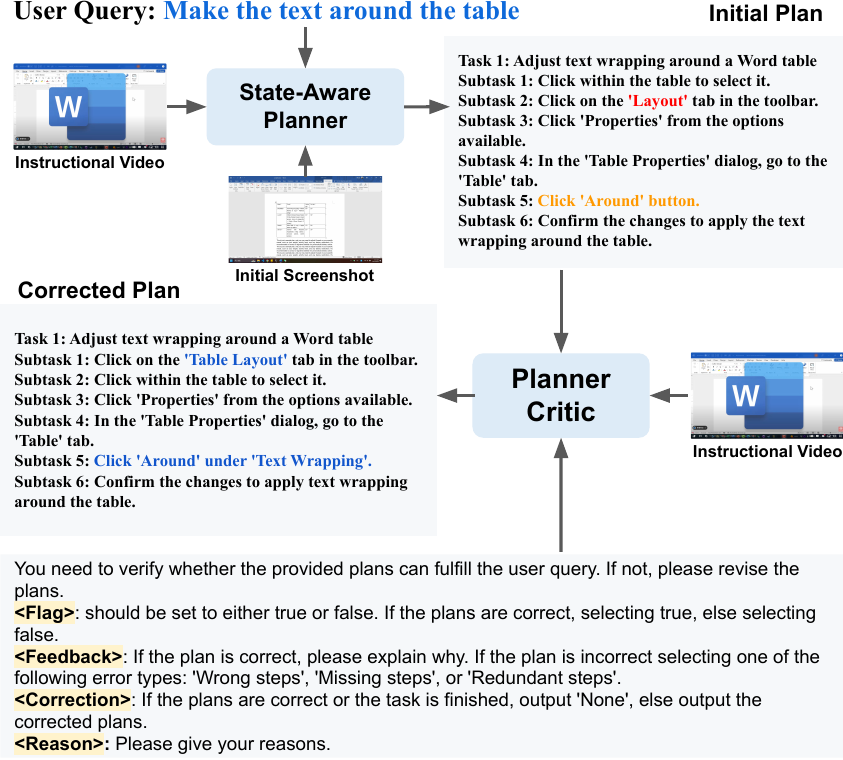}
\caption{\textbf{State-Aware Planner and Planner-Critic.} The Planner generates an initial plan. Then, the Planner-Critic provides necessary corrections.}
\label{fig:plannercritic}

\end{figure}

\noindent\textbf{$\bullet$ Post-Planning Critique:} After the planning phase, a critique module verifies and, if necessary, self-corrects the generated plans to ensure their accuracy.

\noindent\textbf{$\bullet$ Pre-Action Validation:} Before executing each subtask, a validation module determines whether the subtask should be executed. This step is crucial, as the current GUI environment may indicate that the subtask is unnecessary or requires modification to align with the current state.

\noindent\textbf{$\bullet$ Post-Action Evaluation:} After each action execution, a mechanism evaluates whether the action was successfully completed before proceeding to the next subtask.

These critique designs ensure the reliability and adaptability of \agent{} in complex GUI environments.

\subsection{State-Aware Planner} 
The State-Aware Planner processes the instructional video $v$ and the user query $q$ generates an initial plan as shown in the left of Figure~\ref{fig:plannercritic}. We use the speech recognition model Whisper \citep{radford2023robust} to translate the video $v$ into the subtitle and then send it to the MLLM for task planning. The task plan is hierarchically structured as $p=[p_1, p_2, ..., p_N]$ where $p_i$ is a text string describing the $i$-th milestone of the task. Under each $p_i$, there is a list of subtasks $[S^i_1, S^i_2, S^i_N]$, where $S^i_j$ is the $j$-th subtask in the $i$-th milestone. To ensure the produced plans fit the GUI environment, we propose incorporating an initial screenshot $V_0$ to represent the current state. This additional context allows the agent to output plans that align with the actual state. For example, if the instructional video suggests clicking on the ``Layout'' tab in the Word application, but the current state (as indicated by the screenshot) shows that the ``Layout'' tab is already selected, there is no need to perform this action again. By utilizing the visual information from the screenshot, the State-Aware Planner can modify the plans accordingly, rather than strictly following the guidance in the instructional video or the existing knowledge from backbone MLLMs. It also avoids the occlusion issue when not seeing the screenshot.

\subsection{Planner-Critic}
\textbf{Post-Planning Critique.} The goal of the Planner-Critic is to assess the correctness of the initial plans generated by the State-Aware Planner and provide corrections if needed. This module is designed to ensure the accuracy of the plans while leveraging the self-reflection capabilities of MLLMs. As illustrated in Figure~\ref{fig:plannercritic}, for each Initial Plan, the output consists of four components: 

(1) \verb|<Flag>|: Indicates whether the Initial Plan is correct.

(2) \verb|<Feedback>|: Identifies the error type, categorized into one of three options: ``Wrong Steps,'' ``Missing Steps,'' or ``Redundant Steps.''

(3) \verb|<Correction>|: Provide the corrected plans if the Flag indicates that the Initial Plan is incorrect.

(4) \verb|<Reason>|: In addition to giving the corrected plans, we force the model to give the reasons. As previous studies (e.g., CoT~\citep{wei2022chain}, Deepseek-R1~\citep{deepseekr1}) demonstrate that generating reasoning steps along with the answer would enhance the performance.

\subsection{Step-Check}

\begin{figure}[t]
\centering

\includegraphics[width=\linewidth]{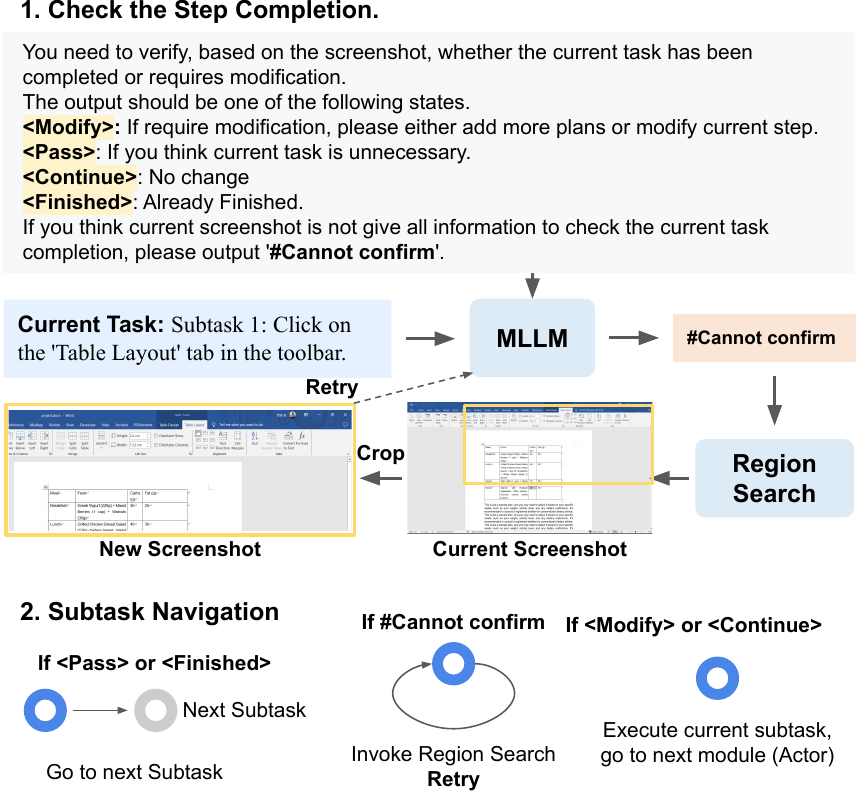}

\caption{\textbf{Step-Check.} This module first checks the step completion status via an MLLM and then navigates to the current task processing.}
\label{fig:stepcheck}

\end{figure}

\textbf{Pre-Action Validation.} After the plan assessment, a navigation mechanism is crucial before sending each subtask $S_t=S_j^i$ at the time step $t$ to the Actor module. To address this, we designed a new module called Step-Check. Through extensive investigation, we discovered that during GUI task testing, perfect execution plans are rarely feasible due to the unpredictable nature of real application environments. Most software retains user preferences (e.g., remember the last configuration of user), meaning that when executing a specific task, the plan $p$ generated by the Planner might not align with the actual state of the software. Therefore, the model must determine whether to proceed with a subtask $S_t$ based on the current state (screenshot: $V_t$, metadata: $M_t$). 

As illustrated in Figure~\ref{fig:stepcheck}, we employ an MLLM to determine whether the current task has been completed or requires modification. We systematically categorize the possible outcomes into: 

(1) \verb|<Modify>|: Indicates that the subtask should be modified or additional subtasks should be added. 

(2) \verb|<Pass>|: Indicates that the current subtask is unnecessary and can be skipped. 

(3) \verb|<Continue>|: Indicates that the subtask is valid and should be executed as planned. 

(4) \verb|<Finished>|: Indicates that the subtask has already been completed and requires no further action.

In cases where the screenshot does not provide sufficient visual information for the MLLM to determine the output, the model outputs \textbf{``$\#$Cannot confirm''}. When this occurs, we design a \textbf{Region Search} module implemented by an LLM. This module takes the corresponding GUI information extracted by the GUI parser and the task description of the current subtask to identify the relevant region. It then crops the region using the generated bounding box as the center coordinate, with the maximum width and height set to half of the original screenshot dimensions (ensure the region is smaller than the original screenshot). The cropped screenshot is subsequently sent to the Step-Check module to regenerate the decision.

\subsection{Actor}

The goal of the Actor is to translate natural language subtask $S_{t}$ into executable code $C_t$. Using an MLLM as the backbone model, the Actor processes metadata $m_t$ and screenshot $V_t$ as GUI context to generate precise executable actions, such as \texttt{click(100, 200)}. Additionally, it leverages the history of previous actions as memory to aid in generating subsequent actions. The generated actions will be executed in the environment, and then the new screenshot $V_{t+1}$ and metadata $m_{t+1}$ will be captured for the next processing.

\subsection{Actor-Critic}

\begin{figure}[t]

\centering
\centerline{\includegraphics[width=0.9\linewidth]{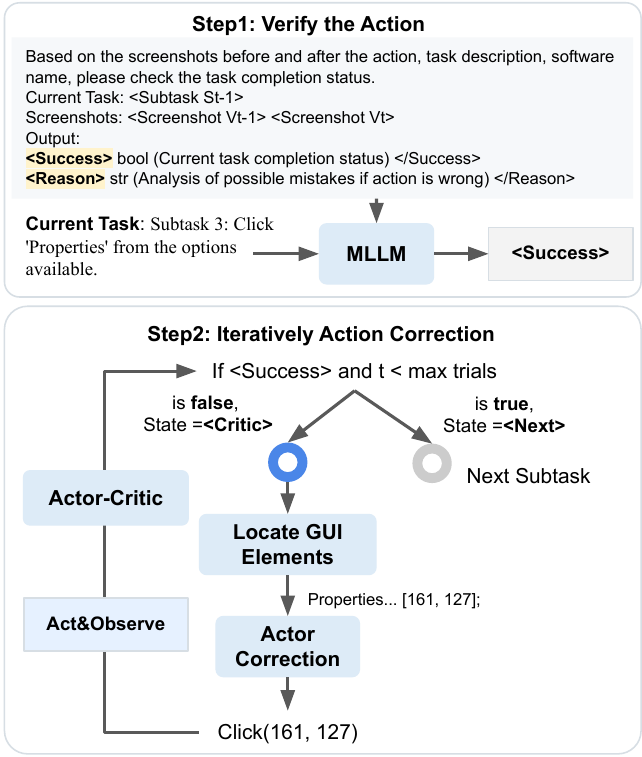}}
\caption{\textbf{Actor-Critic.} This module includes two parts: task verification and task correction. The design follows the \textit{verify-then-correct} mechanism.} 
\label{fig:actorcritic}

\end{figure}

\textbf{Post-Action Evaluation.} After generating an action, the Actor-Critic module evaluates subtask $S_{t-1}$ completion and makes corrections if necessary. As illustrated in Figure \ref{fig:actorcritic}, in the first step, the module implemented by an MLLM compares screenshots \(V_{t-1}\) (before action execution) and \(V_t\) (after execution) while processing each subtask $S_t$ to determine the action correctness. The model outputs a \verb|<Success>| flag to indicate task completion. If the \verb|<Success>| flag is true, the current state $s_t=\verb|<Next>|$. If the \verb|<Success>| flag is false (set $s_t=\verb|<Critic>|$) and the number of trial steps is below the maximum limit, the Actor-Critic module activates the \textbf{Locate GUI Elements} and \textbf{Actor Correction} processes. We introduce the module \textbf{Locate GUI Elements} to identify the relevant GUI elements and regenerate actions using the \textbf{Actor Correction} module. The corrected actions are then executed in the environment, generating updated observations (\(\mathcal{O}_{t}\)) that include new screenshots and metadata for the continued Actor-Critic iteration. The process repeats until the \verb|<Success>| flag is true or the maximum number of trials is reached.

%% file: main.bib
@InProceedings{assistgui,
    author    = {Gao, Difei and Ji, Lei and Bai, Zechen and Ouyang, Mingyu and Li, Peiran and Mao, Dongxing and Wu, Qinchen and Zhang, Weichen and Wang, Peiyi and Guo, Xiangwu and Wang, Hengxu and Zhou, Luowei and Shou, Mike Zheng},
    title     = {AssistGUI: Task-Oriented PC Graphical User Interface Automation},
    booktitle = {Proceedings of the IEEE/CVF Conference on Computer Vision and Pattern Recognition (CVPR)},
    month     = {June},
    year      = {2024},
    pages     = {13289-13298}
}

@inproceedings{webarena,
  title={WebArena: A Realistic Web Environment for Building Autonomous Agents},
  author={Zhou, Shuyan and Xu, Frank F and Zhu, Hao and Zhou, Xuhui and Lo, Robert and Sridhar, Abishek and Cheng, Xianyi and Ou, Tianyue and Bisk, Yonatan and Fried, Daniel and others},
  booktitle={The Twelfth International Conference on Learning Representations}
}

@misc{OSWorld,
      title={OSWorld: Benchmarking Multimodal Agents for Open-Ended Tasks in Real Computer Environments}, 
      author={Tianbao Xie and Danyang Zhang and Jixuan Chen and Xiaochuan Li and Siheng Zhao and Ruisheng Cao and Toh Jing Hua and Zhoujun Cheng and Dongchan Shin and Fangyu Lei and Yitao Liu and Yiheng Xu and Shuyan Zhou and Silvio Savarese and Caiming Xiong and Victor Zhong and Tao Yu},
      year={2024},
      eprint={2404.07972},
      archivePrefix={arXiv},
      primaryClass={cs.AI}
}

@article{webshop,
  title={WebShop: Towards Scalable Real-World Web Interaction with Grounded Language Agents},
  author={Yao, Shunyu and Chen, Howard and Yang, John and Narasimhan, Karthik},
  journal={Advances in neural information processing systems},
  year={2022}
}

@misc{windowsagentarena,
      title={Windows Agent Arena: Evaluating Multi-Modal OS Agents at Scale}, 
      author={Rogerio Bonatti and Dan Zhao and Francesco Bonacci and Dillon Dupont and Sara Abdali and Yinheng Li and Yadong Lu and Justin Wagle and Kazuhito Koishida and Arthur Bucker and Lawrence Jang and Zack Hui},
      year={2024},
      eprint={2409.08264},
      archivePrefix={arXiv},
      primaryClass={cs.AI},
      url={https://arxiv.org/abs/2409.08264}, 
}

@inproceedings{hong2024cogagent,
  title={Cogagent: A visual language model for gui agents},
  author={Hong, Wenyi and Wang, Weihan and Lv, Qingsong and Xu, Jiazheng and Yu, Wenmeng and Ji, Junhui and Wang, Yan and Wang, Zihan and Dong, Yuxiao and Ding, Ming and others},
  booktitle={Proceedings of the IEEE/CVF Conference on Computer Vision and Pattern Recognition},
  pages={14281--14290},
  year={2024}
}

@article{cheng2024seeclick,
  title={Seeclick: Harnessing gui grounding for advanced visual gui agents},
  author={Cheng, Kanzhi and Sun, Qiushi and Chu, Yougang and Xu, Fangzhi and Li, Yantao and Zhang, Jianbing and Wu, Zhiyong},
  journal={arXiv preprint arXiv:2401.10935},
  year={2024}
}

@article{zhang2023appagent,
  title={Appagent: Multimodal agents as smartphone users},
  author={Zhang, Chi and Yang, Zhao and Liu, Jiaxuan and Han, Yucheng and Chen, Xin and Huang, Zebiao and Fu, Bin and Yu, Gang},
  journal={arXiv preprint arXiv:2312.13771},
  year={2023}
}

@inproceedings{he2016deep,
  title={Deep residual learning for image recognition},
  author={He, Kaiming and Zhang, Xiangyu and Ren, Shaoqing and Sun, Jian},
  booktitle={Proceedings of the IEEE conference on computer vision and pattern recognition},
  pages={770--778},
  year={2016}
}

@inproceedings{antol2015vqa,
  title={Vqa: Visual question answering},
  author={Antol, Stanislaw and Agrawal, Aishwarya and Lu, Jiasen and Mitchell, Margaret and Batra, Dhruv and Zitnick, C Lawrence and Parikh, Devi},
  booktitle={Proceedings of the IEEE international conference on computer vision},
  pages={2425--2433},
  year={2015}
}

@article{visualwebarena,
  title={Visualwebarena: Evaluating multimodal agents on realistic visual web tasks},
  author={Koh, Jing Yu and Lo, Robert and Jang, Lawrence and Duvvur, Vikram and Lim, Ming Chong and Huang, Po-Yu and Neubig, Graham and Zhou, Shuyan and Salakhutdinov, Ruslan and Fried, Daniel},
  journal={arXiv preprint arXiv:2401.13649},
  year={2024}
}

@article{lin2024showui,
  title={ShowUI: One Vision-Language-Action Model for GUI Visual Agent},
  author={Lin, Kevin Qinghong and Li, Linjie and Gao, Difei and Yang, Zhengyuan and Wu, Shiwei and Bai, Zechen and Lei, Weixian and Wang, Lijuan and Shou, Mike Zheng},
  journal={arXiv preprint arXiv:2411.17465},
  year={2024}
}

@misc{claude3.5,
  title={Introducing computer use, a new Claude 3.5 Sonnet, and Claude 3.5 Haiku},
  author={Anthropic},
  url={https://www.anthropic.com/news/3-5-models-and-computer-use},
  year={2024}
}

@misc{claude4,
  title={Introducing Claude 4},
  author={Anthropic},
  url={https://www.anthropic.com/news/claude-4},
  year={2025}
}

@inproceedings{radford2023robust,
  title={Robust speech recognition via large-scale weak supervision},
  author={Radford, Alec and Kim, Jong Wook and Xu, Tao and Brockman, Greg and McLeavey, Christine and Sutskever, Ilya},
  booktitle={International conference on machine learning},
  pages={28492--28518},
  year={2023},
  organization={PMLR}
}

@misc{hu2024dawnguiagentpreliminary,
      title={The Dawn of GUI Agent: A Preliminary Case Study with Claude 3.5 Computer Use}, 
      author={Siyuan Hu and Mingyu Ouyang and Difei Gao and Mike Zheng Shou},
      year={2024},
      eprint={2411.10323},
      archivePrefix={arXiv},
      primaryClass={cs.AI},
      url={https://arxiv.org/abs/2411.10323}, 
}

@article{wei2022chain,
  title={Chain-of-thought prompting elicits reasoning in large language models},
  author={Wei, Jason and Wang, Xuezhi and Schuurmans, Dale and Bosma, Maarten and Xia, Fei and Chi, Ed and Le, Quoc V and Zhou, Denny and others},
  journal={Advances in neural information processing systems},
  volume={35},
  pages={24824--24837},
  year={2022}
}

@misc{deepseekr1,
      title={DeepSeek-R1: Incentivizing Reasoning Capability in LLMs via Reinforcement Learning}, 
      author={DeepSeek-AI and Daya Guo and Dejian Yang and Haowei Zhang and Junxiao Song and Ruoyu Zhang and Runxin Xu and Qihao Zhu and Shirong Ma and Peiyi Wang and Xiao Bi and Xiaokang Zhang and Xingkai Yu and Yu Wu and Z. F. Wu and Zhibin Gou and Zhihong Shao and Zhuoshu Li and Ziyi Gao and Aixin Liu and Bing Xue and Bingxuan Wang and Bochao Wu and Bei Feng and Chengda Lu and Chenggang Zhao and Chengqi Deng and Chenyu Zhang and Chong Ruan and Damai Dai and Deli Chen and Dongjie Ji and Erhang Li and Fangyun Lin and Fucong Dai and Fuli Luo and Guangbo Hao and Guanting Chen and Guowei Li and H. Zhang and Han Bao and Hanwei Xu and Haocheng Wang and Honghui Ding and Huajian Xin and Huazuo Gao and Hui Qu and Hui Li and Jianzhong Guo and Jiashi Li and Jiawei Wang and Jingchang Chen and Jingyang Yuan and Junjie Qiu and Junlong Li and J. L. Cai and Jiaqi Ni and Jian Liang and Jin Chen and Kai Dong and Kai Hu and Kaige Gao and Kang Guan and Kexin Huang and Kuai Yu and Lean Wang and Lecong Zhang and Liang Zhao and Litong Wang and Liyue Zhang and Lei Xu and Leyi Xia and Mingchuan Zhang and Minghua Zhang and Minghui Tang and Meng Li and Miaojun Wang and Mingming Li and Ning Tian and Panpan Huang and Peng Zhang and Qiancheng Wang and Qinyu Chen and Qiushi Du and Ruiqi Ge and Ruisong Zhang and Ruizhe Pan and Runji Wang and R. J. Chen and R. L. Jin and Ruyi Chen and Shanghao Lu and Shangyan Zhou and Shanhuang Chen and Shengfeng Ye and Shiyu Wang and Shuiping Yu and Shunfeng Zhou and Shuting Pan and S. S. Li and Shuang Zhou and Shaoqing Wu and Shengfeng Ye and Tao Yun and Tian Pei and Tianyu Sun and T. Wang and Wangding Zeng and Wanjia Zhao and Wen Liu and Wenfeng Liang and Wenjun Gao and Wenqin Yu and Wentao Zhang and W. L. Xiao and Wei An and Xiaodong Liu and Xiaohan Wang and Xiaokang Chen and Xiaotao Nie and Xin Cheng and Xin Liu and Xin Xie and Xingchao Liu and Xinyu Yang and Xinyuan Li and Xuecheng Su and Xuheng Lin and X. Q. Li and Xiangyue Jin and Xiaojin Shen and Xiaosha Chen and Xiaowen Sun and Xiaoxiang Wang and Xinnan Song and Xinyi Zhou and Xianzu Wang and Xinxia Shan and Y. K. Li and Y. Q. Wang and Y. X. Wei and Yang Zhang and Yanhong Xu and Yao Li and Yao Zhao and Yaofeng Sun and Yaohui Wang and Yi Yu and Yichao Zhang and Yifan Shi and Yiliang Xiong and Ying He and Yishi Piao and Yisong Wang and Yixuan Tan and Yiyang Ma and Yiyuan Liu and Yongqiang Guo and Yuan Ou and Yuduan Wang and Yue Gong and Yuheng Zou and Yujia He and Yunfan Xiong and Yuxiang Luo and Yuxiang You and Yuxuan Liu and Yuyang Zhou and Y. X. Zhu and Yanhong Xu and Yanping Huang and Yaohui Li and Yi Zheng and Yuchen Zhu and Yunxian Ma and Ying Tang and Yukun Zha and Yuting Yan and Z. Z. Ren and Zehui Ren and Zhangli Sha and Zhe Fu and Zhean Xu and Zhenda Xie and Zhengyan Zhang and Zhewen Hao and Zhicheng Ma and Zhigang Yan and Zhiyu Wu and Zihui Gu and Zijia Zhu and Zijun Liu and Zilin Li and Ziwei Xie and Ziyang Song and Zizheng Pan and Zhen Huang and Zhipeng Xu and Zhongyu Zhang and Zhen Zhang},
      year={2025},
      eprint={2501.12948},
      archivePrefix={arXiv},
      primaryClass={cs.CL},
      url={https://arxiv.org/abs/2501.12948}, 
}

@misc{androidworld,
      title={AndroidWorld: A Dynamic Benchmarking Environment for Autonomous Agents}, 
      author={Christopher Rawles and Sarah Clinckemaillie and Yifan Chang and Jonathan Waltz and Gabrielle Lau and Marybeth Fair and Alice Li and William Bishop and Wei Li and Folawiyo Campbell-Ajala and Daniel Toyama and Robert Berry and Divya Tyamagundlu and Timothy Lillicrap and Oriana Riva},
      year={2024},
      eprint={2405.14573},
      archivePrefix={arXiv},
      primaryClass={cs.AI},
      url={https://arxiv.org/abs/2405.14573}, 
}

@misc{autowebglm,
      title={AutoWebGLM: A Large Language Model-based Web Navigating Agent}, 
      author={Hanyu Lai and Xiao Liu and Iat Long Iong and Shuntian Yao and Yuxuan Chen and Pengbo Shen and Hao Yu and Hanchen Zhang and Xiaohan Zhang and Yuxiao Dong and Jie Tang},
      year={2024},
      eprint={2404.03648},
      archivePrefix={arXiv},
      primaryClass={cs.CL},
      url={https://arxiv.org/abs/2404.03648}, 
}

@misc{osatlas,
      title={OS-ATLAS: A Foundation Action Model for Generalist GUI Agents}, 
      author={Zhiyong Wu and Zhenyu Wu and Fangzhi Xu and Yian Wang and Qiushi Sun and Chengyou Jia and Kanzhi Cheng and Zichen Ding and Liheng Chen and Paul Pu Liang and Yu Qiao},
      year={2024},
      eprint={2410.23218},
      archivePrefix={arXiv},
      primaryClass={cs.CL},
      url={https://arxiv.org/abs/2410.23218}, 
}

@inproceedings{zheng2024seeact,
  title={GPT-4V(ision) is a Generalist Web Agent, if Grounded},
  author={Boyuan Zheng and Boyu Gou and Jihyung Kil and Huan Sun and Yu Su},
  booktitle={Forty-first International Conference on Machine Learning},
  year={2024},
  url={https://openreview.net/forum?id=piecKJ2DlB},
}

@inproceedings{webvoyager,
    title = "{W}eb{V}oyager: Building an End-to-End Web Agent with Large Multimodal Models",
    author = "He, Hongliang  and
      Yao, Wenlin  and
      Ma, Kaixin  and
      Yu, Wenhao  and
      Dai, Yong  and
      Zhang, Hongming  and
      Lan, Zhenzhong  and
      Yu, Dong",
    editor = "Ku, Lun-Wei  and
      Martins, Andre  and
      Srikumar, Vivek",
    booktitle = "Proceedings of the 62nd Annual Meeting of the Association for Computational Linguistics (Volume 1: Long Papers)",
    month = aug,
    year = "2024",
    address = "Bangkok, Thailand",
    publisher = "Association for Computational Linguistics",
    url = "https://aclanthology.org/2024.acl-long.371/",
    doi = "10.18653/v1/2024.acl-long.371",
    pages = "6864--6890"
}

@article{wang2024mobile,
  title={Mobile-Agent: Autonomous Multi-Modal Mobile Device Agent with Visual Perception},
  author={Wang, Junyang and Xu, Haiyang and Ye, Jiabo and Yan, Ming and Shen, Weizhou and Zhang, Ji and Huang, Fei and Sang, Jitao},
  journal={arXiv preprint arXiv:2401.16158},
  year={2024}
}

@inproceedings{VQAv2,
  title={Making the v in vqa matter: Elevating the role of image understanding in visual question answering},
  author={Goyal, Yash and Khot, Tejas and Summers-Stay, Douglas and Batra, Dhruv and Parikh, Devi},
  booktitle={CVPR},
  year={2017}
}

@inproceedings{AutoDroid,
author = {Wen, Hao and Li, Yuanchun and Liu, Guohong and Zhao, Shanhui and Yu, Tao and Li, Toby Jia-Jun and Jiang, Shiqi and Liu, Yunhao and Zhang, Yaqin and Liu, Yunxin},
title = {AutoDroid: LLM-powered Task Automation in Android},
year = {2024},
isbn = {9798400704895},
publisher = {Association for Computing Machinery},
address = {New York, NY, USA},
url = {https://doi.org/10.1145/3636534.3649379},
doi = {10.1145/3636534.3649379},
booktitle = {Proceedings of the 30th Annual International Conference on Mobile Computing and Networking},
pages = {543–557},
numpages = {15},
location = {Washington D.C., DC, USA},
series = {ACM MobiCom '24}
}

@inproceedings{
zheng2025agentstudio,
title={AgentStudio: A Toolkit for Building General Virtual Agents},
author={Longtao Zheng and Zhiyuan Huang and Zhenghai Xue and Xinrun Wang and Bo An and Shuicheng YAN},
booktitle={The Thirteenth International Conference on Learning Representations},
year={2025},
url={https://openreview.net/forum?id=axUf8BOjnH}
}

@misc{agents2,
      title={Agent S2: A Compositional Generalist-Specialist Framework for Computer Use Agents}, 
      author={Saaket Agashe and Kyle Wong and Vincent Tu and Jiachen Yang and Ang Li and Xin Eric Wang},
      year={2025},
      eprint={2504.00906},
      archivePrefix={arXiv},
      primaryClass={cs.AI},
      url={https://arxiv.org/abs/2504.00906}, 
}

@inproceedings{
agents1,
title={Agent S: An Open Agentic Framework that Uses Computers Like a Human},
author={Saaket Agashe and Jiuzhou Han and Shuyu Gan and Jiachen Yang and Ang Li and Xin Eric Wang},
booktitle={The Thirteenth International Conference on Learning Representations},
year={2025},
url={https://openreview.net/forum?id=lIVRgt4nLv}
}

@misc{gpt-5,
      title={OpenAI GPT-5 System Card}, 
      author={Aaditya Singh and Adam Fry and Adam Perelman and Adam Tart and Adi Ganesh and Ahmed El-Kishky and Aidan McLaughlin and Aiden Low and AJ Ostrow and Akhila Ananthram and Akshay Nathan and Alan Luo and Alec Helyar and Aleksander Madry and Aleksandr Efremov and Aleksandra Spyra and Alex Baker-Whitcomb and Alex Beutel and Alex Karpenko and Alex Makelov and Alex Neitz and Alex Wei and Alexandra Barr and Alexandre Kirchmeyer and Alexey Ivanov and Alexi Christakis and Alistair Gillespie and Allison Tam and Ally Bennett and Alvin Wan and Alyssa Huang and Amy McDonald Sandjideh and Amy Yang and Ananya Kumar and Andre Saraiva and Andrea Vallone and Andrei Gheorghe and Andres Garcia Garcia and Andrew Braunstein and Andrew Liu and Andrew Schmidt and Andrey Mereskin and Andrey Mishchenko and Andy Applebaum and Andy Rogerson and Ann Rajan and Annie Wei and Anoop Kotha and Anubha Srivastava and Anushree Agrawal and Arun Vijayvergiya and Ashley Tyra and Ashvin Nair and Avi Nayak and Ben Eggers and Bessie Ji and Beth Hoover and Bill Chen and Blair Chen and Boaz Barak and Borys Minaiev and Botao Hao and Bowen Baker and Brad Lightcap and Brandon McKinzie and Brandon Wang and Brendan Quinn and Brian Fioca and Brian Hsu and Brian Yang and Brian Yu and Brian Zhang and Brittany Brenner and Callie Riggins Zetino and Cameron Raymond and Camillo Lugaresi and Carolina Paz and Cary Hudson and Cedric Whitney and Chak Li and Charles Chen and Charlotte Cole and Chelsea Voss and Chen Ding and Chen Shen and Chengdu Huang and Chris Colby and Chris Hallacy and Chris Koch and Chris Lu and Christina Kaplan and Christina Kim and CJ Minott-Henriques and Cliff Frey and Cody Yu and Coley Czarnecki and Colin Reid and Colin Wei and Cory Decareaux and Cristina Scheau and Cyril Zhang and Cyrus Forbes and Da Tang and Dakota Goldberg and Dan Roberts and Dana Palmie and Daniel Kappler and Daniel Levine and Daniel Wright and Dave Leo and David Lin and David Robinson and Declan Grabb and Derek Chen and Derek Lim and Derek Salama and Dibya Bhattacharjee and Dimitris Tsipras and Dinghua Li and Dingli Yu and DJ Strouse and Drew Williams and Dylan Hunn and Ed Bayes and Edwin Arbus and Ekin Akyurek and Elaine Ya Le and Elana Widmann and Eli Yani and Elizabeth Proehl and Enis Sert and Enoch Cheung and Eri Schwartz and Eric Han and Eric Jiang and Eric Mitchell and Eric Sigler and Eric Wallace and Erik Ritter and Erin Kavanaugh and Evan Mays and Evgenii Nikishin and Fangyuan Li and Felipe Petroski Such and Filipe de Avila Belbute Peres and Filippo Raso and Florent Bekerman and Foivos Tsimpourlas and Fotis Chantzis and Francis Song and Francis Zhang and Gaby Raila and Garrett McGrath and Gary Briggs and Gary Yang and Giambattista Parascandolo and Gildas Chabot and Grace Kim and Grace Zhao and Gregory Valiant and Guillaume Leclerc and Hadi Salman and Hanson Wang and Hao Sheng and Haoming Jiang and Haoyu Wang and Haozhun Jin and Harshit Sikchi and Heather Schmidt and Henry Aspegren and Honglin Chen and Huida Qiu and Hunter Lightman and Ian Covert and Ian Kivlichan and Ian Silber and Ian Sohl and Ibrahim Hammoud and Ignasi Clavera and Ikai Lan and Ilge Akkaya and Ilya Kostrikov and Irina Kofman and Isak Etinger and Ishaan Singal and Jackie Hehir and Jacob Huh and Jacqueline Pan and Jake Wilczynski and Jakub Pachocki and James Lee and James Quinn and Jamie Kiros and Janvi Kalra and Jasmyn Samaroo and Jason Wang and Jason Wolfe and Jay Chen and Jay Wang and Jean Harb and Jeffrey Han and Jeffrey Wang and Jennifer Zhao and Jeremy Chen and Jerene Yang and Jerry Tworek and Jesse Chand and Jessica Landon and Jessica Liang and Ji Lin and Jiancheng Liu and Jianfeng Wang and Jie Tang and Jihan Yin and Joanne Jang and Joel Morris and Joey Flynn and Johannes Ferstad and Johannes Heidecke and John Fishbein and John Hallman and Jonah Grant and Jonathan Chien and Jonathan Gordon and Jongsoo Park and Jordan Liss and Jos Kraaijeveld and Joseph Guay and Joseph Mo and Josh Lawson and Josh McGrath and Joshua Vendrow and Joy Jiao and Julian Lee and Julie Steele and Julie Wang and Junhua Mao and Kai Chen and Kai Hayashi and Kai Xiao and Kamyar Salahi and Kan Wu and Karan Sekhri and Karan Sharma and Karan Singhal and Karen Li and Kenny Nguyen and Keren Gu-Lemberg and Kevin King and Kevin Liu and Kevin Stone and Kevin Yu and Kristen Ying and Kristian Georgiev and Kristie Lim and Kushal Tirumala and Kyle Miller and Lama Ahmad and Larry Lv and Laura Clare and Laurance Fauconnet and Lauren Itow and Lauren Yang and Laurentia Romaniuk and Leah Anise and Lee Byron and Leher Pathak and Leon Maksin and Leyan Lo and Leyton Ho and Li Jing and Liang Wu and Liang Xiong and Lien Mamitsuka and Lin Yang and Lindsay McCallum and Lindsey Held and Liz Bourgeois and Logan Engstrom and Lorenz Kuhn and Louis Feuvrier and Lu Zhang and Lucas Switzer and Lukas Kondraciuk and Lukasz Kaiser and Manas Joglekar and Mandeep Singh and Mandip Shah and Manuka Stratta and Marcus Williams and Mark Chen and Mark Sun and Marselus Cayton and Martin Li and Marvin Zhang and Marwan Aljubeh and Matt Nichols and Matthew Haines and Max Schwarzer and Mayank Gupta and Meghan Shah and Melody Huang and Meng Dong and Mengqing Wang and Mia Glaese and Micah Carroll and Michael Lampe and Michael Malek and Michael Sharman and Michael Zhang and Michele Wang and Michelle Pokrass and Mihai Florian and Mikhail Pavlov and Miles Wang and Ming Chen and Mingxuan Wang and Minnia Feng and Mo Bavarian and Molly Lin and Moose Abdool and Mostafa Rohaninejad and Nacho Soto and Natalie Staudacher and Natan LaFontaine and Nathan Marwell and Nelson Liu and Nick Preston and Nick Turley and Nicklas Ansman and Nicole Blades and Nikil Pancha and Nikita Mikhaylin and Niko Felix and Nikunj Handa and Nishant Rai and Nitish Keskar and Noam Brown and Ofir Nachum and Oleg Boiko and Oleg Murk and Olivia Watkins and Oona Gleeson and Pamela Mishkin and Patryk Lesiewicz and Paul Baltescu and Pavel Belov and Peter Zhokhov and Philip Pronin and Phillip Guo and Phoebe Thacker and Qi Liu and Qiming Yuan and Qinghua Liu and Rachel Dias and Rachel Puckett and Rahul Arora and Ravi Teja Mullapudi and Raz Gaon and Reah Miyara and Rennie Song and Rishabh Aggarwal and RJ Marsan and Robel Yemiru and Robert Xiong and Rohan Kshirsagar and Rohan Nuttall and Roman Tsiupa and Ronen Eldan and Rose Wang and Roshan James and Roy Ziv and Rui Shu and Ruslan Nigmatullin and Saachi Jain and Saam Talaie and Sam Altman and Sam Arnesen and Sam Toizer and Sam Toyer and Samuel Miserendino and Sandhini Agarwal and Sarah Yoo and Savannah Heon and Scott Ethersmith and Sean Grove and Sean Taylor and Sebastien Bubeck and Sever Banesiu and Shaokyi Amdo and Shengjia Zhao and Sherwin Wu and Shibani Santurkar and Shiyu Zhao and Shraman Ray Chaudhuri and Shreyas Krishnaswamy and Shuaiqi and Xia and Shuyang Cheng and Shyamal Anadkat and Simón Posada Fishman and Simon Tobin and Siyuan Fu and Somay Jain and Song Mei and Sonya Egoian and Spencer Kim and Spug Golden and SQ Mah and Steph Lin and Stephen Imm and Steve Sharpe and Steve Yadlowsky and Sulman Choudhry and Sungwon Eum and Suvansh Sanjeev and Tabarak Khan and Tal Stramer and Tao Wang and Tao Xin and Tarun Gogineni and Taya Christianson and Ted Sanders and Tejal Patwardhan and Thomas Degry and Thomas Shadwell and Tianfu Fu and Tianshi Gao and Timur Garipov and Tina Sriskandarajah and Toki Sherbakov and Tomer Kaftan and Tomo Hiratsuka and Tongzhou Wang and Tony Song and Tony Zhao and Troy Peterson and Val Kharitonov and Victoria Chernova and Vineet Kosaraju and Vishal Kuo and Vitchyr Pong and Vivek Verma and Vlad Petrov and Wanning Jiang and Weixing Zhang and Wenda Zhou and Wenlei Xie and Wenting Zhan and Wes McCabe and Will DePue and Will Ellsworth and Wulfie Bain and Wyatt Thompson and Xiangning Chen and Xiangyu Qi and Xin Xiang and Xinwei Shi and Yann Dubois and Yaodong Yu and Yara Khakbaz and Yifan Wu and Yilei Qian and Yin Tat Lee and Yinbo Chen and Yizhen Zhang and Yizhong Xiong and Yonglong Tian and Young Cha and Yu Bai and Yu Yang and Yuan Yuan and Yuanzhi Li and Yufeng Zhang and Yuguang Yang and Yujia Jin and Yun Jiang and Yunyun Wang and Yushi Wang and Yutian Liu and Zach Stubenvoll and Zehao Dou and Zheng Wu and Zhigang Wang},
      year={2025},
      eprint={2601.03267},
      archivePrefix={arXiv},
      primaryClass={cs.CL},
      url={https://arxiv.org/abs/2601.03267}, 
}

@misc{uitars,
      title={UI-TARS: Pioneering Automated GUI Interaction with Native Agents}, 
      author={Yujia Qin and Yining Ye and Junjie Fang and Haoming Wang and Shihao Liang and Shizuo Tian and Junda Zhang and Jiahao Li and Yunxin Li and Shijue Huang and Wanjun Zhong and Kuanye Li and Jiale Yang and Yu Miao and Woyu Lin and Longxiang Liu and Xu Jiang and Qianli Ma and Jingyu Li and Xiaojun Xiao and Kai Cai and Chuang Li and Yaowei Zheng and Chaolin Jin and Chen Li and Xiao Zhou and Minchao Wang and Haoli Chen and Zhaojian Li and Haihua Yang and Haifeng Liu and Feng Lin and Tao Peng and Xin Liu and Guang Shi},
      year={2025},
      eprint={2501.12326},
      archivePrefix={arXiv},
      primaryClass={cs.AI},
      url={https://arxiv.org/abs/2501.12326}, 
}

@article{UIExplorer,
  title={Toward Autonomous UI Exploration: The UIExplorer Benchmark},
  author={Nica, Andrei Cristian and Kudlu Shanbhogue, Akshaya Vishnu and Shah, Harshil and Cambray, Aleix and Berariu, Tudor and Maystre, Lucas and Barber, David},
  journal={arXiv preprint arXiv:2506.17779},
  year={2025}
}

@misc{guiworld2025,
  title={GUI-World: A Dataset for GUI-Oriented Multimodal LLM-Based Evaluation},
  author={Chen, Dongping and others},
  year={2025},
  howpublished={\url{https://gui-world.github.io}}
}

@inproceedings{
    yang2025guirobust,
    title={GUI-Robust: A Comprehensive Dataset for Testing GUI Agent Robustness in Real-World Anomalies},
    author={Jingqi Yang and Zhilong Song and Jiawei Chen and Mingli Song and Sheng Zhou and Linjun Sun and Xiaogang Ouyang and Chun Chen and Can Wang},
    booktitle={NeurIPS Datasets and Benchmarks Track},
    year={2025},
    url={https://openreview.net/forum?id=22gw3kITCd},
}

@article{xu2024aguvis,
  title={Aguvis: Unified Pure Vision Agents for Autonomous GUI Interaction},
  author={Yiheng Xu and Zekun Wang and Junli Wang and Dunjie Lu and Tianbao Xie and Amrita Saha and Doyen Sahoo and Tao Yu and Caiming Xiong},
  year={2024},
  url={https://arxiv.org/abs/2412.04454}
}

@inproceedings{gou2024uground,
title={Navigating the Digital World as Humans Do: Universal Visual Grounding for {GUI} Agents},
author={Boyu Gou and Ruohan Wang and Boyuan Zheng and Yanan Xie and Cheng Chang and Yiheng Shu and Huan Sun and Yu Su},
booktitle={The Thirteenth International Conference on Learning Representations},
year={2025},
url={https://openreview.net/forum?id=kxnoqaisCT}
}

@inproceedings{
xu2025agenttrek,
title={AgentTrek: Agent Trajectory Synthesis via Guiding Replay with Web Tutorials},
author={Yiheng Xu and Dunjie Lu and Zhennan Shen and Junli Wang and Zekun Wang and Yuchen Mao and Caiming Xiong and Tao Yu},
booktitle={The Thirteenth International Conference on Learning Representations},
year={2025},
url={https://openreview.net/forum?id=EEgYUccwsV}
}

@inproceedings{
yuan2025segui,
title={{SE}-{GUI}: Enhancing Visual Grounding for {GUI} Agents via Self-Evolutionary Reinforcement Learning},
author={Xinbin Yuan and Jian Zhang and Kaixin Li and Zhuoxuan Cai and Lujian Yao and Jie Chen and Enguang Wang and Qibin Hou and Jinwei Chen and Peng-Tao Jiang and Bo Li},
booktitle={The Thirty-ninth Annual Conference on Neural Information Processing Systems},
year={2025},
url={https://openreview.net/forum?id=IbzDaIDyt6}
}
